%% file: paper.tex
\documentclass[]{bytedance_seed}

\usepackage[toc,page,header]{appendix}

\usepackage{minitoc}

\usepackage{amsthm}
\usepackage{caption}

\newtheorem{proposition}{Proposition}[section]

\input{math_commands.tex}

\usepackage{subcaption}

\usepackage{graphicx}
\usepackage{booktabs}
\usepackage{hyperref}
\usepackage{url}
\usepackage{multirow}

\usepackage{dblfloatfix}

\usepackage{cuted}
\usepackage{caption}
\usepackage{subcaption}

\usepackage{wrapfig}
\usepackage{booktabs}

\usepackage{algorithm}
\usepackage{algpseudocode}
\usepackage{xcolor}
 \usepackage[table]{xcolor}

\algrenewcommand{\algorithmicrequire}{\textbf{Input:}}
\algrenewcommand{\algorithmicensure}{\textbf{Output:}}

\algrenewcommand{\algorithmiccomment}[1]{%
    \hfill$\triangleright$\ {\color{gray}\textnormal{#1}}%
}

\newenvironment{ruledalgorithm}[1]{%
    \par\noindent
    \refstepcounter{algorithm}%
    \hrule height 0.8pt
    \vspace{3pt}
    {\normalsize\bfseries
        Algorithm~\thealgorithm\quad #1\par
    }
    \vspace{3pt}
    \hrule height 0.4pt
    \vspace{3pt}
}{%
    \vspace{3pt}
    \hrule height 0.4pt
    \par
}

\title{Aligning One-Step Generative Models with Reward-Weighted Transport Distillation}

\author[1]{Austin Wang}
\author[1,2]{Ziheng Cheng}
\author[1]{Lexing Ying}

\affiliation[1]{ByteDance Seed}
\affiliation[2]{UC Berkeley}

\abstract{
One-step generators enable high-quality visual generation with a single network
evaluation, but their post-training is difficult: general implicit generators
provide neither tractable likelihoods nor denoising trajectories, and many
rewards are non-differentiable. We introduce \textit{Reward-Weighted Transport
Distillation} (RWTD), a post-training method that requires only generated samples and scalar reward evaluations. Rather than aligning solely to the conventional reward-tilted reference distribution, RWTD constructs an adaptive target that mixes separately tilted current and reference distributions. The current component incorporates improvements discovered during training, while the reference component anchors the target to the pretrained generator. RWTD realizes this target through feature-space optimal transport and fixed-point regression.
Theoretical analysis shows that the fixed-point distributions of RWTD interpolate between off-policy reward tilting of the reference and on-policy tilting of the current model, providing a principled approach to balancing reward adaptation with retention of prior knowledge. Empirically, RWTD substantially improves the GenEval score of the one-step SANA Sprint 1.6B backbone from 0.73 to 0.80, while separate preference alignment experiments demonstrate strong cross-reward generalization that yields balanced improvements and preservation of compositional capabilities.
}

\correspondence{Austin Wang at \email{austinwang@bytedance.com}}

\checkdata[Code and Model Checkpoints]{\href{https://github.com/austin-k-wang/Reward-Weighted-Transport-Distillation/tree/main}{github.com/austin-k-wang/Reward-Weighted-Transport-Distillation}}

\begin{document}
\maketitle

%不需要目录就注释掉 注意目录不要和第一页放在一块 要有\newpage
%\newpage
%\tableofcontents
%\newpage

\input{sections/introduction}
\input{sections/relatedwork}
\input{sections/background}
\input{sections/approach}
\input{sections/theory}
\input{sections/experiments}

% \clearpage

\bibliographystyle{plainnat}
\bibliography{main}

\clearpage

\beginappendix

\input{sections/appendix}

\end{document}

%% file: math_commands.tex
\usepackage{amsmath,amsfonts,bm}

\def\eqref#1{equation~\ref{#1}}
\def\1{\bm{1}}

\def\eps{{\epsilon}}

\def\vone{{\bm{1}}}

\def\va{{\bm{a}}}
\def\vb{{\bm{b}}}

\def\vs{{\bm{s}}}
\def\vt{{\bm{t}}}
\def\vu{{\bm{u}}}
\def\vv{{\bm{v}}}
\def\vw{{\bm{w}}}
\def\vx{{\bm{x}}}
\def\vy{{\bm{y}}}
\def\vz{{\bm{z}}}

\def\mC{{\bm{C}}}

\def\mK{{\bm{K}}}

\def\mP{{\bm{P}}}

\DeclareMathAlphabet{\mathsfit}{\encodingdefault}{\sfdefault}{m}{sl}
\SetMathAlphabet{\mathsfit}{bold}{\encodingdefault}{\sfdefault}{bx}{n}

\newcommand{\E}{\mathbb{E}}
\newcommand{\Ls}{\mathcal{L}}
\newcommand{\R}{\mathbb{R}}

\newcommand{\lr}{\alpha}

\newcommand{\softmax}{\mathrm{softmax}}

\DeclareMathOperator*{\argmax}{arg\,max}
\DeclareMathOperator*{\argmin}{arg\,min}

%% file: sections/introduction.tex
\section{Introduction}
\label{sec:introduction}
% Budget: 1.25 pages.

Diffusion~\citep{ho2020denoising,song2020score,karras2022elucidating} and flow-based models~\citep{lipman2022flow,albergo2025stochastic} achieve high-fidelity visual synthesis but typically require many sequential network evaluations. Distillation methods—including consistency models~\citep{song2023consistency,luo2023latent}, adversarial diffusion distillation~\citep{sauer2024adversarial}, and distribution matching distillation~\citep{yin2024one,yin2024improved}—instead compress multi-step teachers into few- or one-step generators. More recently, objectives based on drifting~\citep{deng2026generative}, Wasserstein gradient flows~\citep{han2026one}, mean flows~\citep{geng2026mean, geng2026improved}, and inductive moment matching~\citep{zhou2025inductive} have enabled one-step generators to be trained directly. As these models improve, developing effective post-training methods for general one-step generators becomes increasingly important.

Existing alignment methods do not yet provide a general interface. Diffusion reinforcement learning and preference optimization commonly require differentiable rewards~\citep{clark2024directly}, likelihood surrogates~\citep{wallace2024diffusion, wang2026beyond, xue2025advantage}, or access to multi-step denoising or flow trajectories~\citep{black2024training,fan2023dpok,liu2026flow}. Recent methods support one- and few-step generators but remain tied to diffusion dynamics~\citep{wu2026diff, luo2024diff}, rely on high-variance reward-gradient estimation~\citep{lee2026aligning}, or use specialized preference-field constructions~\citep{jiang2026drifting}. We instead seek a method that is agnostic to the generator's pre-training procedure and compatible with black-box rewards.

To this end, we introduce Reward-Weighted Transport Distillation (RWTD),
whose central ingredient is an adaptive alignment target that separately
tilts and mixes current and reference distributions. The current component
incorporates improvements discovered during training, while the reference
component anchors the target to pretrained modes. Feature-space optimal transport
then determines how current samples should move toward this target, and the
resulting assignments are amortized through feature regression using only
scalar reward evaluations. To our knowledge, RWTD is the first one-step
alignment method to explicitly target a mixture of separately normalized
reward-tilted current and reference distributions. Our theoretical analysis characterizes the resulting fixed points: off-policy targeting recovers the conventional reward-tilted reference, while on-policy tilting concentrates on reward-maximizing modes. Intermediate mixtures define a distinct fixed-point family with a reward-dependent rational amplification controlled by the reference mass, providing a principled trade-off between reward adaptation and reference anchoring.

{
\setlength{\textfloatsep}{8pt plus 1pt minus 2pt}
\begin{figure}[t]
    \centering
    \includegraphics[width=0.62\linewidth]{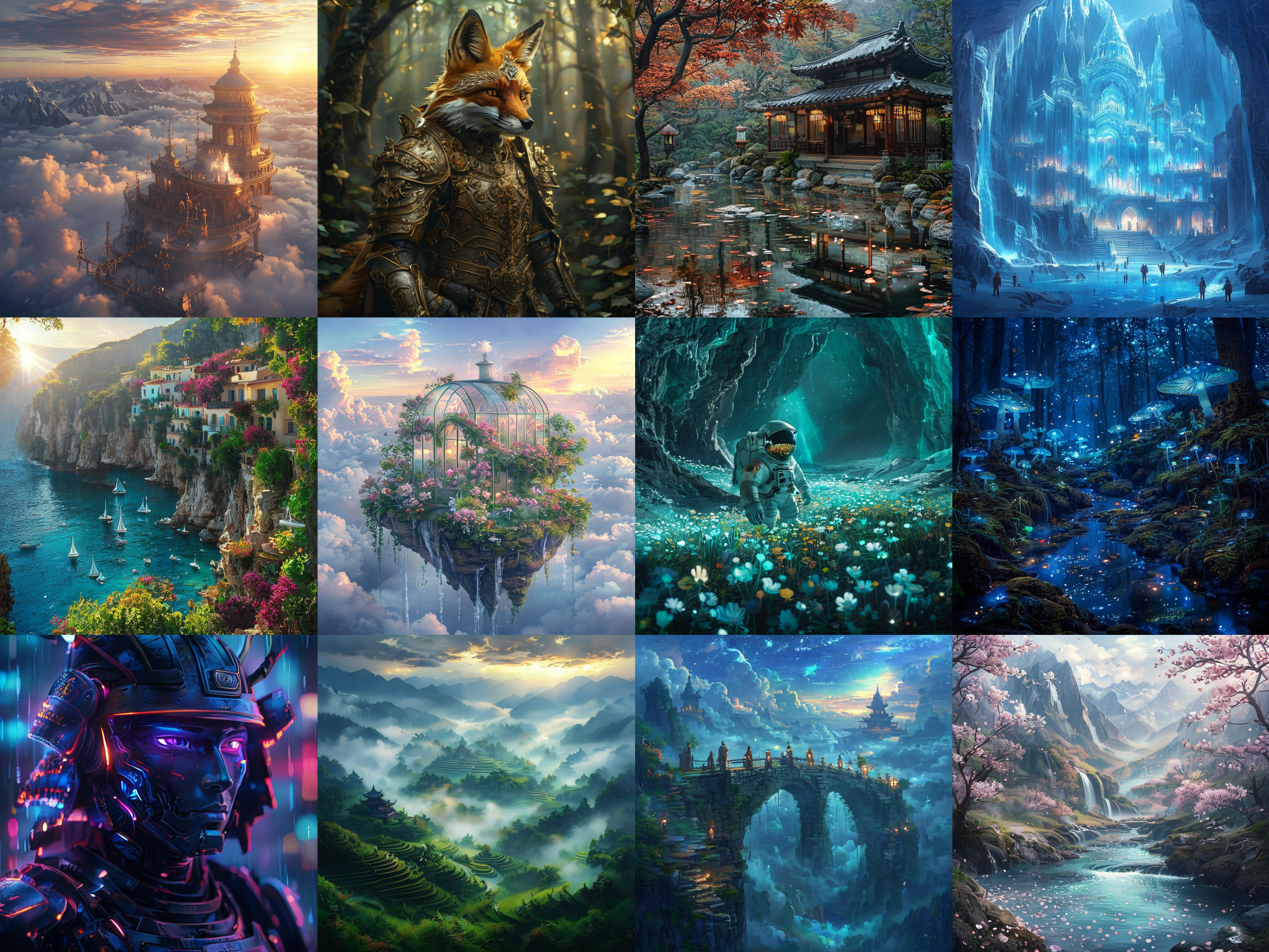}
    \caption{Images generated in one step by SANA Sprint 1.6B after RWTD post-training.}
    \label{fig:sana-samples}
\end{figure}
}

Using non-differentiable compositional feedback, RWTD improves the GenEval score of one-step SANA Sprint 1.6B~\citep{chen2025sana} from \textbf{0.73 to 0.80}, establishing state-of-the-art performance among SANA Sprint-based one-step generators and surpassing multi-step models of comparable scale. When trained with HPSv2~\citep{wu2023human}, RWTD also improves the optimized reward while generalizing across held-out alignment metrics. It produces more balanced gains and better preserves semantic and compositional capabilities, while gradient-based alternatives actively degrade performance on out-of-domain tasks. Detailed ablations isolate how the specific design choices of RWTD improve overall performance and data efficiency. Our main contributions are:
\begin{itemize}
    \setlength{\itemsep}{1pt}
    \setlength{\parsep}{0pt}
    \setlength{\parskip}{0pt}
    \setlength{\topsep}{2pt}
    \setlength{\partopsep}{0pt}

    \item \textbf{A new adaptive alignment target.}
    RWTD replaces reference-only tilting with a mixture of reward-tilted current and reference distributions, realized through feature-space OT using only samples and scalar (potentially black-box) reward evaluations.

    \item \textbf{Interpretable theoretical analysis.}
    We characterize the mixed reward-tilting fixed points, revealing a reward-dependent rational amplification beyond standard reference tilting.

    \item \textbf{Strong compositional and preference alignment.}
    RWTD achieves SOTA GenEval performance at SANA Sprint 1.6B scale. Separate preference alignment experiments show strong cross-reward generalization.
\end{itemize}

%% file: sections/relatedwork.tex
\section{Related Work}
\label{sec:related-work}
% Budget: 0.80 pages.

\subsection{One-Step Generative Models}

One-step generators are trained typically by distilling multi-step teachers through consistency, adversarial, or distribution-matching objectives~\citep{song2023consistency,luo2023latent,sauer2024adversarial,yin2024one,yin2024improved,liu2023instaflow}, or through native objectives based on drifting, Wasserstein gradient flows, mean flows, and inductive moment matching~\citep{deng2026generative,han2026one,geng2026mean,geng2026improved,zhou2025inductive}. RWTD addresses the complementary problem of aligning an existing one-step generator with a new scalar reward.

\subsection{Reward Alignment of Visual Generators}

Visual generators have been aligned through direct reward backpropagation \citep{prabhudesai2023aligning,clark2024directly}, reinforcement learning \citep{black2024training,fan2023dpok,liu2026flow, zheng2026diffusionnft, choi2026rethinking, xue2025advantage}, and preference optimization \citep{wallace2024diffusion,wang2026beyond, lu2025inpo, lu2025smoothed, han2026discrete}. These approaches can substantially improve target rewards, but generally assume differentiable rewards, tractable likelihood surrogates, or access to multi-step denoising or flow trajectories. Complementary inference-time steering methods adapt generation through reward- or likelihood-guided sampling, search, or derivative-free posterior inference, but leave the generator unchanged and require additional sampling computation~\citep{singhal2025general,uehara2025inference,huang2026guide,bansal2024universal,yeh2025training,zheng2025blade, wang2025ensemble, zheng2024ensemble}. We focus on post-training alignment that amortizes reward improvement into the generator while preserving one-step inference. 

Recent methods align few- and one-step generators but retain important restrictions. DI++, DI*, and DIDR~\citep{luo2024diff+,luo2024diff, wu2026diff} are diffusion-specific and require reward gradients, while FAV~\citep{lee2026aligning} requires kernel density estimation (KDE) and reward gradients. DrPO~\citep{jiang2026drifting} avoids gradients by constructing a rank-based dipole preference field from generated candidates. RWTD instead introduces a different alignment target—an adaptive mixture of reward-tilted current and reference distributions—and realizes it through feature-space OT, without requiring reward gradients, KDE, or access to a multi-step diffusion teacher.

\subsection{Optimal Transport for One-Step Generative Modeling}

Optimal transport (OT) provides a geometric framework for moving probability mass between distributions, while entropic regularization enables efficient approximation through the Sinkhorn algorithm \citep{cuturi2013sinkhorn,gabriel2019computational}. Recently, W-Flow~\citep{han2026one} trains a native one-step generator by defining a Wasserstein gradient flow from a reference distribution toward the data distribution and amortizing its evolution into a neural network. RWTD instead addresses post-training, using transport to realize reward-driven distributional updates for an existing generator.

%% file: sections/background.tex
\section{Background and Problem Setting}
\label{sec:background}
% Budget: 0.65 pages.

\subsection{Black-Box Alignment of One-Step Generators}
% Define G_\theta(z,c), the frozen reference generator, scalar reward access,
% and the unavailable likelihood, score, trajectory, and reward gradient.

Let $G_\theta$ be a conditional one-step generator that maps a latent variable $\vz \sim p_z$ and condition $c \sim p_{\mathrm{prompt}}$ directly to a sample $\vx = G_\theta(\vz,c),$ inducing the implicit conditional distribution $p_\theta(\vx\mid c)$. We also assume access to a frozen reference generator $G_{\mathrm{ref}}$, with distribution $p_{\mathrm{ref}}(\vx\mid c)$, and a scalar reward function $r(\vx,c)$. Our objective is to optimize the reward scores of our model's samples while preserving the quality, diversity, and conditional fidelity of the reference generator.

We consider a black-box alignment setting in which $p_\theta$ and $p_{\mathrm{ref}}$ need not have tractable likelihoods, and $r$ need not be differentiable. In particular, we do not assume access to a diffusion trajectory, score function, likelihood ratio, or reward gradient. 
% Although our primary application is one-step visual generation, this formulation applies to any implicit generator satisfying these sampling assumptions.

A standard operation for converting rewards into a target distribution is
exponential reward tilting~\citep{wallace2024diffusion,uehara2024understanding,
han2026discrete,fan2023dpok}. For any conditional base distribution
$p(\vx\mid c)$ and inverse temperature $\beta>0$, define

\begin{equation}
\begin{aligned}
Z_p(c)
&=
\E_{\vx\sim p(\cdot\mid c)}
\left[\exp\!\left(\beta r(\vx,c)\right)\right], \\
\mathcal{T}_{\beta,r}[p](\vx\mid c)
&=
\frac{
p(\vx\mid c)\exp\!\left(\beta r(\vx,c)\right)
}{
Z_p(c)
}.
\end{aligned}
\label{eq:tilt-operator}
\end{equation}

The inverse temperature $\beta$ controls alignment strength: as
$\beta\rightarrow0$, the tilted distribution approaches $p$, while larger
values concentrate mass on high-reward regions. Reward tilting has the
variational characterization~\citep{donsker1975asymptotic}

\begin{equation}
\mathcal{T}_{\beta,r}[p](\cdot\mid c)
=
\argmax_q
\left\{
\E_q[r(\vx,c)]
-
\frac{1}{\beta}
\mathbb{D}_{\mathrm{KL}}
\!\left(q(\cdot\mid c)\,\Vert\,p(\cdot\mid c)\right)
\right\}.
\label{eq:tilt-variational}
\end{equation}

Thus, reward tilting increases expected reward while penalizing deviation
from the base distribution. RWTD builds on this canonical alignment
primitive by mixing tilted current and reference distributions.

\subsection{Entropic Optimal Transport}
\label{sec:background_ot}

Consider empirical measures $\mu=\sum_{i=1}^{N}a_i\delta_{\vu_i}$ and
$\nu=\sum_{j=1}^{M}b_j\delta_{\vv_j}$, where $\va$ and $\vb$ are probability
vectors. A transport coupling between $\mu$ and $\nu$ is a nonnegative matrix whose row and column sums match these prescribed weights. The set of admissible couplings is
$\Pi(\va,\vb)=\{\mP\in\R_+^{N\times M}:
\mP\vone_M=\va,\ \mP^\top\vone_N=\vb\}$.
Given pairwise costs $C_{ij}$ between $\vu_i$ and $\vv_j$, entropic OT computes

\begin{equation}
    \mP_\eps
    =
    \argmin_{\mP\in\Pi(\va,\vb)}
    \left\{
        \langle\mP,\mC\rangle
        +
        \eps\sum_{i,j}
        P_{ij}\bigl(\log P_{ij}-1\bigr)
    \right\}.
    \label{eq:entropic-ot}
\end{equation}

The regularization $\eps>0$ smooths the coupling, while $\eps\rightarrow0$
recovers unregularized OT under standard conditions. Each row of the coupling represents how much mass is assigned from a source particle to each target particle. The Sinkhorn algorithm efficiently computes this coupling by iteratively rescaling the kernel $\exp(-\mC/\eps)$ to match the prescribed marginals
\citep{cuturi2013sinkhorn}.

%% file: sections/approach.tex
\section{Reward-Weighted Transport Distillation}
\label{sec:method}
% Budget: 2.00 pages.
% Reserve space for Figure 1: the RWTD pipeline.

\subsection{Mixed Reward-Tilted Particle Target}
\label{sec:mixed-target}

The central design choice in RWTD is the target distribution. Unlike canonical alignment to a fixed reward-tilted reference, RWTD constructs a target that evolves with the current generator while retaining an explicit reference anchor. Given the current generator distribution $p_\theta$ and reference distribution $p_\text{ref}$, we define the mixed reward-tilted target as
\begin{equation} \label{eq:mixed-target}
    Q_{\beta, \rho}[p_\theta] = (1 - \rho) \mathcal{T}_{\beta,r}[p_\theta] + \rho \mathcal{T}_{\beta,r}[p_\text{ref}],
\end{equation}

where $0 \leq \rho \leq 1$ is a scalar hyperparameter controlling the mass allocated to the reward-tilted reference distribution. Intuitively, $\rho$ trades off on-policy and off-policy tilting behavior. The on-policy component $\mathcal{T}_{\beta,r}[p_\theta]$ exploits high-reward regions discovered by the current generator, but evolves as the generator is updated. The off-policy component $\mathcal{T}_{\beta,r}[p_{\mathrm{ref}}]$ instead provides a stationary source of high-reward samples from the pretrained model, continually reintroducing modes that may have been lost by the current generator. The endpoints $\rho=0$ and $\rho=1$ correspond to purely on-policy and purely off-policy targets, respectively.

The mixed target in Equation~\ref{eq:mixed-target} can be approximated with particles. For a conditioning input $c$, we draw independent latent variables and generate current and reference particles as

\begin{equation}
    \begin{aligned}
        \vz_i &\sim p_z,
        & \vx_i &= G_\theta(\vz_i,c),
        && i=1,\ldots,N, \\
        \vz'_j &\sim p_z,
        & \vy_j &= G_{\mathrm{ref}}(\vz'_j,c),
        && j=1,\ldots,M.
    \end{aligned}
    \label{eq:sampled-particles}
\end{equation}

We then form separately normalized reward weights for the current and reference particles:

\begin{equation}
    \begin{aligned}
        w_i^\theta
        &=
        \frac{
            \exp\!\left(\beta r(\vx_i,c)\right)
        }{
            \sum_{\ell=1}^{N}
            \exp\!\left(\beta r(\vx_\ell,c)\right)
        },
        && i=1,\ldots,N, \\
        w_j^{\mathrm{ref}}
        &=
        \frac{
            \exp\!\left(\beta r(\vy_j,c)\right)
        }{
            \sum_{\ell=1}^{M}
            \exp\!\left(\beta r(\vy_\ell,c)\right)
        },
        && j=1,\ldots,M.
    \end{aligned}
    \label{eq:particle-reward-weights}
\end{equation}

The resulting empirical target measure and uniformly weighted source measure are

\begin{equation}
    \begin{aligned}
        \widehat{Q}_{\beta,\rho}[p_\theta]
        &=
        (1-\rho)
        \sum_{i=1}^{N}
        w_i^\theta\delta_{\vx_i}
        +
        \rho
        \sum_{j=1}^{M}
        w_j^{\mathrm{ref}}\delta_{\vy_j},
        \\
        \widehat{p}_\theta
        &=
        \frac{1}{N}
        \sum_{i=1}^{N}
        \delta_{\vx_i}.
    \end{aligned}
    \label{eq:empirical-particle-measures}
\end{equation}

Separate normalization ensures that the current and reference particles receive total masses $1-\rho$ and $\rho$, respectively. This construction requires only generated samples and scalar reward evaluations; neither generator likelihoods nor reward gradients are needed. We treat $\widehat{p}_\theta$ as the source measure and $\widehat{Q}_{\beta,\rho}[p_\theta]$ as the target measure, and construct a feature-space optimal-transport coupling between them in the next section.

\subsection{Feature-Space Optimal Transport}
\label{sec:feature-ot}

We construct a transport coupling from the uniformly weighted current distribution to the mixed reward-weighted target in Equation~\ref{eq:empirical-particle-measures}. Because Euclidean distance in pixel space poorly reflects perceptual and semantic similarity, we instead perform transport in the representation space of a frozen visual encoder $\phi$, following standard practice~\citep{deng2026generative, han2026one}.

For each current particle, let $\vu_i=\phi(\vx_i)$ denote its source feature. We concatenate the encoded current and reference particles into the target features by setting $\vv_i=\phi(\vx_i)$ for $i=1,\ldots,N$ and $\vv_{N+j}=\phi(\vy_j)$ for $j=1,\ldots,M$. Collecting the reward weights into vectors $\vw^\theta=(w_1^\theta,\ldots,w_N^\theta)^\top$ and $\vw^{\mathrm{ref}}=(w_1^{\mathrm{ref}},\ldots,w_M^{\mathrm{ref}})^\top$, the source and target marginals are $\va=\vone_N/N$ and $\vb=[(1-\rho)\vw^\theta;\rho\vw^{\mathrm{ref}}]$, respectively. We define the feature-space cost by $C_{ik}=\|\vu_i-\vv_k\|_2^2$, yielding $\mC\in\R^{N\times(N+M)}$. Using the Sinkhorn algorithm, we solve the entropic OT problem described in Section~\ref{sec:background_ot}:

\begin{equation}
    \mP_\eps
    =
    \operatorname{Sinkhorn}_{\eps}
    \left(\mC;\va,\vb\right)
    \in\Pi(\va,\vb).
    \label{eq:rwtd-transport-plan}
\end{equation}

The resulting plan satisfies $\mP_\eps\vone_{N+M}=\va$ and $\mP_\eps^\top\vone_N=\vb$. Each row specifies how the mass of one current particle is distributed across the target particles. The coupling geometrically realizes the mixed target without reward gradients or diffusion-specific quantities. In the next subsection, we convert the conditional assignments into sample-specific regression targets. 

\subsection{Partial Transport Distillation}
\label{sec:transport-distillation}

Each row of $\mP_\eps$ defines a conditional distribution over mixed target particles after normalization by its mass $a_i=1/N$. RWTD partially moves each source feature toward the conditional mean:

\begin{equation}
    \widetilde{\vu}_i
    =
    (1-\eta)\vu_i
    +
    \eta
    \underbrace{
        \frac{1}{a_i}
        \sum_{k=1}^{N+M}
        P_{\eps,ik}\vv_k
    }_{\bar{\vu}_i},
    \qquad 0<\eta\leq1,
    \label{eq:barycentric-target}
    % \label{eq:partial-transport-target}
\end{equation}

where $\bar{\vu}_i$ is the barycentric destination and $\eta$ is the transport step size. Setting $\eta=1$ uses the full destination, while smaller values produce more conservative updates. Since practical couplings are estimated from finite minibatches and amortized by regression, partial steps are used to limit sensitivity to coupling noise and abrupt distribution shifts.

RWTD amortizes these particle-level transport targets into the generator by regressing the features of samples generated from the original latents toward their detached targets:

\begin{equation}
    \mathcal{L}_{\mathrm{RWTD}}(\theta)
    =
    \frac{1}{N}
    \sum_{i=1}^{N}
    \left\|
        \phi\!\left(G_\theta(\vz_i,c)\right)
        -
        \operatorname{sg}\!\left(
            \widetilde{\vu}_i
        \right)
    \right\|_2^2,
    \label{eq:rwtd-objective}
\end{equation}

where $\operatorname{sg}(\cdot)$ denotes the stop-gradient operator. The encoder $\phi$ remains frozen, but gradients propagate through $\phi(G_\theta(\vz_i,c))$ to update the generator. No gradients are taken through the reward weights, transport coupling, or target construction. Detailed algorithm pseudocode is provided in Algorithms \ref{alg:rwtd} and \ref{alg:sinkhorn}.

Because the targets are recomputed from the current generator after each update, this objective constitutes a form of fixed-point regression: RWTD alternates between constructing transport targets from the current distribution and distilling those targets back into the generator. This motivates a fixed-point interpretation, made precise in the next section.

%% file: sections/theory.tex
\section{Theoretical Analysis}
\label{sec:theory}

We analyze an idealized population-level version of RWTD that isolates the distributional dynamics induced by reward tilting and mixing. We fix a condition $c$ and suppress it from the notation.

\subsection{Fixed Points of Mixed Reward Tilting}
\label{sec:theory-fixed-points}

Consider the idealized full-update recursion

\begin{equation}
    p_{t+1}
    =
    Q_{\beta,\rho}[p_t]
    =
    (1-\rho)\mathcal{T}_{\beta,r}[p_t]
    +
    \rho\mathcal{T}_{\beta,r}[p_{\mathrm{ref}}].
    \label{eq:population-recursion}
\end{equation}

We characterize fixed points of this operator and contrast it with the
canonical reward-tilted reference obtained at $\rho=1$.

\begin{proposition}[Characterization of mixed reward-tilting fixed points]
\label{prop:mixed-fixed-point}
Let $0<\rho\leq1$ and consider fixed points satisfying
$p_\infty\ll p_{\mathrm{ref}}$. Define
\[
    H(\lambda)
    =
    \E_{\vx\sim \mathcal{T}_{\beta,r}[p_{\mathrm{ref}}]}
    \left[
        \frac{1}{1-\lambda \exp(\beta r(\vx))}
    \right]
\]
on the set of nonnegative $\lambda$ for which the denominator is
positive $\mathcal{T}_{\beta,r}[p_{\mathrm{ref}}]$-almost everywhere and the expectation is
finite. Fixed points are in one-to-one correspondence with
solutions of
\begin{equation}
    H(\lambda)=\frac{1}{\rho}.
    \label{eq:lambda-characterization}
\end{equation}
Any such solution is unique and the corresponding
fixed point is
\begin{equation}
    p_\infty(\vx)
    =
    \frac{\rho}{Z_{\mathrm{ref}}}
    \frac{
        p_{\mathrm{ref}}(\vx)\exp(\beta r(\vx))
    }{
        1-\lambda\exp(\beta r(\vx))
    }.
\end{equation}
In particular, when $\rho=1$, the unique solution is $\lambda=0$ and
$p_\infty=\mathcal{T}_{\beta,r}[p_{\mathrm{ref}}]$, recovering the standard reward-tilted reference.
\end{proposition}

The proof and further analysis are deferred to
Sections~\ref{app:mixed-fixed-point} and~\ref{app:rho-analysis}.
Mixed targeting is not equivalent to interpolating model parameters or merely
changing the reward temperature. At $\rho=1$, the operator recovers the
conventional reward-tilted reference distribution, whereas for
$0<\rho<1$, any fixed point contains the additional rational factor
$\left(1-\lambda \exp({\beta r(\vx))}\right)^{-1}$, defining a distinct family
of distributions that preferentially amplifies high-reward regions. Decreasing $\rho$ increases $\lambda$ and sharpens this amplification. 

The purely on-policy endpoint $\rho=0$ must be treated separately because it has no unique fixed point. In this case,

\begin{equation}
    p_t(\vx)
    =
    \frac{
        p_0(\vx)\exp\!\left(t\beta r(\vx)\right)
    }{
        \E_{\vx'\sim p_0}
        \left[
            \exp\!\left(t\beta r(\vx')\right)
        \right]
    }.
    \label{eq:on-policy-iterates}
\end{equation}

which concentrates on the reward-maximizing regions in the support of $p_0$ as $t$ increases. Thus, we view $\rho$ as controlling the trade-off between reference anchoring and recursive reward sharpening: larger values keep the fixed point closer to the reward-tilted reference distribution, whereas smaller values give greater influence to the on-policy component and increasingly concentrate mass on high-reward modes.

\subsection{Fixed-Point Invariance under Partial Transport}
\label{sec:theory-partial-transport}

We next study whether the transport step size $\eta$ changes the fixed points of the idealized population dynamics. Assume that $p$ and $Q_{\beta,\rho}[p]$ are probability measures on $\mathbb{R}^d$ with finite second moments, with $p$ absolutely continuous, and let $T_p$ be the exact quadratic-cost optimal transport map from $p$ to $Q_{\beta,\rho}[p]$, such that $(T_p)_\#p=Q_{\beta,\rho}[p]$. The corresponding RWTD partial transport update is

\begin{equation}
    \mathcal{F}_\eta[p]
    =
    \left(
        (1-\eta)\operatorname{Id}
        +
        \eta T_p
    \right)_\#p,
    \qquad
    0<\eta\leq1,
    \label{eq:partial-population-update}
\end{equation}

where $_\#$ denotes the pushforward of a distribution. This is the population analogue of the partial targets in Equation~\ref{eq:barycentric-target}.

\begin{proposition}[Fixed-point invariance]
\label{prop:partial-fixed-point-invariance}
Under the assumptions above, for every $0<\eta\leq1$,

\begin{equation}
    \mathcal{F}_\eta[p]=p
    \quad\Longleftrightarrow\quad
    Q_{\beta,\rho}[p]=p.
    \label{eq:partial-fixed-point-equivalence}
\end{equation}

Consequently, the set of fixed points is independent of the transport step size $\eta$.
\end{proposition}

The proof is deferred to Section \ref{app:partial-fixed-point-invariance}. Thus, if any nonzero partial step leaves $p$ unchanged, then the full target must already equal $p$. Conversely, if $p=Q_{\beta,\rho}[p]$, the optimal transport map is the identity almost everywhere, so every partial step also leaves $p$ unchanged. The invariance of the idealized fixed point to the transport step size motivates partial transport as a mechanism for stabilizing training through conservative updates while preserving the fixed-point set. 

We emphasize that the theoretical results in this section operate under idealized conditions and therefore provide intuition about the behavior of our method, rather than strict guarantees. Practical RWTD instead uses finite minibatches, entropic OT, and barycentric projections to approximate the idealized dynamics.

%% file: sections/experiments.tex
\section{Experiments}
\label{sec:experiments}
% Budget: 2.35 pages.
% Main-text assets: GenEval table, cross-reward table, and method ablation.

\subsection{Experimental Setup}
\label{sec:experimental-setup}

\begin{figure}
    \centering
    \includegraphics[width=0.8\linewidth]{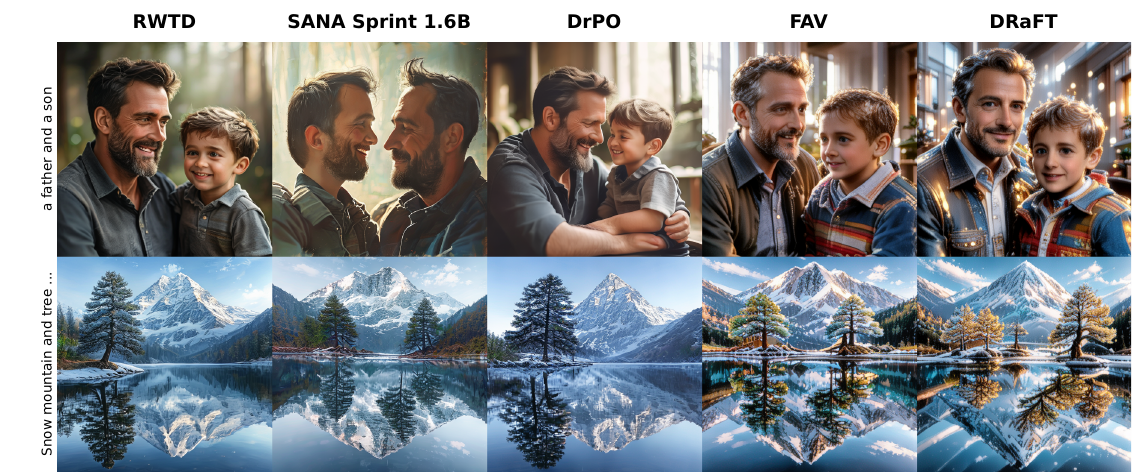}
    \caption{Qualitative comparison of models post-trained with HPSv2. Gradient-based methods FAV and DRaFT exhibit pronounced stylization consistent with reward overoptimization, while RWTD largely preserves realism. For the second prompt, RWTD produces a geometrically coherent lake reflection, while baselines introduce spurious trees and other inconsistencies in the reflected scene.}
    \label{fig:sana-figure}
\end{figure}

\textbf{Dataset and Models} Our experiments used two one-step generation backbones: SDXL Turbo~\citep{sauer2024adversarial, podell2024sdxl} and SANA Sprint 1.6B~\citep{chen2025sana}. For GenEval, we used training prompts generated from the official GenEval codebase (MIT License), filtering prompts to ensure no overlap with the evaluation set, following established setup~\citep{liu2026flow, zheng2026diffusionnft}. For preference alignment experiments, we trained with the HPSv2~\citep{wu2023human} reward model and prompts from the Pick-a-Pic v2~\citep{kirstain2023pick} dataset (MIT License). We use the DINOv2~\citep{oquab2023dinov2} model as the frozen feature encoder during SANA Sprint 1.6B training, and MAE ~\citep{he2022masked} features during SDXL Turbo training.  

\textbf{Baselines} We compare against recent one-step alignment methods that operate directly on the one-step generator without requiring access to a multi-step diffusion teacher or denoising trajectory. These include DrPO~\citep{jiang2026drifting}, which constructs a dipole field toward preferred feature targets, and FAV~\citep{lee2026aligning}, which uses Stein variational gradient descent~\citep{liu2016stein} and kernel density estimation to move samples toward the reward-tilted reference. For preference alignment, we additionally include DRaFT~\citep{clark2024directly}, a direct reward-backpropagation baseline. All baselines are trained with at least the computational budget of RWTD to ensure fair comparison.

\textbf{Evaluation} We use the official GenEval codebase for evaluation of compositional correctness, sampling four images per prompt and reporting average metrics. For preference alignment evaluation, we use the PartiPrompts~\citep{yu2022scaling} dataset, generating five images per prompt and reporting average PickScore~\cite{kirstain2023pick}, HPSv2~\citep{wu2023human}, Aesthetics~\citep{schuhmann2022aesthetic}, CLIP~\citep{radford2021learning}, and ImageReward~\citep{xu2023imagereward}. We also report overall GenEval scores on models post-trained with HPSv2 to measure out-of-domain generalization. In general, random seeds are standardized to ensure reproducibility and fair comparison, and all evaluation protocols follow established setups~\citep{ghosh2023geneval, liu2026flow, wang2026beyond}.

\textbf{Implementation Details} We finetune LoRA adapters ~\citep{hu2021lora} for both generator backbones. SANA Sprint 1.6B training is run on 8 GPUs while SDXL Turbo training uses 4 GPUs. Full hyperparameter configurations, discussion of computational cost, and extended experimental details are deferred to Sections \ref{sec:hyperparameters} \ref{sec:computational-cost}, and \ref{sec:data-and-models}, respectively. Additional hyperparameter ablation and diversity experiments are provided in Section ~\ref{sec:ablations}. 

\subsection{Main Experimental Results}
\label{sec:geneval}

\definecolor{rwtdblue}{RGB}{230,242,255}

\begin{table*}[t]
    \centering

    % ---------------------------------------------------------
    % SDXL results
    % ---------------------------------------------------------
    \begin{minipage}[t]{0.485\textwidth}
        \centering
        \textbf{(a) SDXL Turbo Comparisons}
        \vspace{3pt}

        \resizebox{\linewidth}{!}{%
        \begin{tabular}{clcccccccc}
            \toprule
            & Method
            & NFEs
            & Overall
            & Position
            & Counting
            & \shortstack{Color\\attr.}
            & \shortstack{Single\\obj.}
            & Colors
            & \shortstack{Two\\obj.} \\
            \midrule

            & SDXL
            & 50
            & 0.55 & 0.11 & 0.43 & 0.21
            & 0.98 & 0.88 & 0.71 \\

            \multirow{-2}{*}{%
                % \rotatebox[origin=c]{90}{\textbf{Multi-step}}%
            }
            & LAIR-SDXL
            & 50
            & 0.59 & \textbf{0.14} & 0.40
            & \textbf{0.28}
            & \textbf{1.00}
            & \textbf{0.91}
            & \textbf{0.83} \\

            \midrule

            & SDXL-Turbo
            & 1
            & 0.55 & 0.09 & 0.48 & 0.20
            & 0.99 & 0.87 & 0.68 \\

            & DrPO
            & 1
            & 0.58 & 0.09 & 0.58 & 0.26
            & \textbf{1.00} & 0.87 & 0.67 \\

            \rowcolor{rwtdblue}
            \cellcolor{white}
            \multirow{-3}{*}{%
                % \rotatebox[origin=c]{90}{\textbf{One-step}}%
            }
            & \textbf{RWTD}
            & 1
            & \textbf{0.61}
            & \textbf{0.14}
            & \textbf{0.63}
            & 0.25
            & \textbf{1.00}
            & 0.89
            & 0.75 \\
            \bottomrule
        \end{tabular}%
        }
    \end{minipage}
    \hfill
    % ---------------------------------------------------------
    % SANA results
    % ---------------------------------------------------------
    \begin{minipage}[t]{0.505\textwidth}
        \centering
        \textbf{(b) SANA Sprint 1.6B Comparisons}
        \vspace{3pt}

        \resizebox{\linewidth}{!}{%
        \begin{tabular}{clcccccccc}
            \toprule
            & Method
            & NFEs
            & Overall
            & Position
            & Counting
            & \shortstack{Color\\attr.}
            & \shortstack{Single\\obj.}
            & Colors
            & \shortstack{Two\\obj.} \\
            \midrule

            & FLUX.1 Dev
            & 50
            & 0.66
            & 0.22 & \textbf{0.74} & 0.45 & 0.98 & 0.79 & 0.81 \\

            & Playground v3
            & --
            & 0.76
            & 0.50 & 0.72 & 0.54 & 0.99 & 0.82 & \textbf{0.95} \\

            % \multirow{-2}{*}{%
            %     % \rotatebox[origin=c]{90}{\textbf{Multi-step}}%
            % }
            & SD3.5-L
            & 50
            & 0.71
            & 0.47 & 0.73 & 0.34 & 0.98 & 0.83 & 0.89 \\

            & SANA Sprint 4-step
            & 4
            & 0.75 
            & 0.57 & 0.59 & 0.54 & \textbf{1.00} & \textbf{0.91} & 0.90 \\

            \midrule

            & SANA Sprint 1.6B
            & 1
            & 0.73 & 0.54 & 0.59 & 0.51
            & 0.99 & 0.89 & 0.88 \\

            & DrPO
            & 1
            & 0.75 & 0.59 & 0.59 & 0.54
            & 0.99 & {0.90} & 0.91 \\

            & FAV$^\dagger$
            & 1
            & 0.73 & 0.53 & 0.60 & 0.50
            & 0.99 & 0.89 & 0.89 \\

            \rowcolor{rwtdblue}
            \cellcolor{white}
            \multirow{-4}{*}{%
                % \rotatebox[origin=c]{90}{\textbf{One-step}}%
            }
            & \textbf{RWTD}
            & 1
            & \textbf{0.80}
            & \textbf{0.73}
            & {0.61}
            & \textbf{0.60}
            & \textbf{1.00}
            & {0.90}
            & \textbf{0.95} \\
            \bottomrule
        \end{tabular}%
        }
    \end{minipage}

    \caption{
        GenEval results for SDXL Turbo comparisons (left) and SANA Sprint
        comparisons (right). Horizontal rules separate
        multi-step diffusion/flow models from one-step generators. $\dagger$: FAV uses zeroth order gradient estimation.
    }
    \label{tab:geneval-results}
    
\end{table*}

\textbf{Black-box GenEval Optimization} Table~\ref{tab:geneval-results} presents our main GenEval results. RWTD improves SDXL Turbo from $0.55$ to $0.61$, outperforming DrPO ($0.58$) and even the preference-tuned SDXL variant LAIR-SDXL~\citep{wang2026beyond} (50 NFEs). On SANA Sprint 1.6B, RWTD achieves $0.80$, substantially improving the $0.73$ base model and surpassing powerful multi-step models such as Playground v3~\citep{liu2024playground} and FLUX.1 Dev~\citep{blackforestlabs2024flux}. The largest gains occur on challenging categories such as Position and Color Attribution, demonstrating effective optimization of difficult tasks from non-differentiable scalar feedback. In contrast, DrPO and FAV provide only marginal or no improvement on this backbone, reaching $0.75$ and $0.73$, respectively.

\definecolor{rwtdblue}{RGB}{230,242,255}
\definecolor{groupgray}{RGB}{242,242,242}

\begin{wraptable}{r}{0.62\columnwidth}
    \vspace{-0.8em}
    \centering
    \scriptsize
    \setlength{\tabcolsep}{3pt}
    \renewcommand{\arraystretch}{1.08}

    \resizebox{\linewidth}{!}{%
    \begin{tabular}{lcccccc}
        \toprule
        Method
        & \shortstack{PS\\$\uparrow$}
        & \shortstack{HPSv2\\$\uparrow$ (in-domain)}
        & \shortstack{CLIP\\$\uparrow$}
        & \shortstack{Aes.\\$\uparrow$}
        & \shortstack{IR\\$\uparrow$}
        & \shortstack{GenEval\\$\uparrow$ (OOD)} \\
        \midrule

        \rowcolor{groupgray}
        \multicolumn{7}{l}{\textbf{Base model}} \\

        SANA Sprint 1.6B
        & 22.75
        & 30.31
        & 0.2753
        & 6.565
        & 1.148
        & 0.73 \\

        \midrule
        \rowcolor{groupgray}
        \multicolumn{7}{l}{\textbf{Gradient-free}} \\

        \rowcolor{rwtdblue}
        \textbf{RWTD (ours)}
        & \textbf{23.04}
        & 32.45
        & \textbf{0.2761}
        & 6.800
        & \textbf{1.358}
        & \textbf{0.75} \\

        DrPO
        & 22.90
        & 32.04
        & 0.2706
        & 6.787
        & 1.291
        & 0.72 \\

        \midrule
        \rowcolor{groupgray}
        \multicolumn{7}{l}{\textbf{Gradient-based}} \\

        DRaFT
        & 22.76
        & \textbf{37.02} 
        & 0.2573
        & \textbf{7.252} 
        & 1.312
        & 0.62 \\

        FAV
        & 22.93
        & 35.74
        & 0.2718
        & 6.949
        & 1.322
        & 0.72 \\

        \bottomrule
    \end{tabular}%
    }

    \caption{
        PartiPrompt evaluation after HPSv2 post-training.
    }
    \label{tab:cross-reward-generalization}
    \vspace{-0.8em}
\end{wraptable}

\textbf{Preference Alignment and Cross-Reward Generalization} We additionally present preference-alignment experiments to measure cross-reward generalization. We train on prompts from Pick-a-Pic using HPSv2 as the reward and evaluate preference metrics on PartiPrompts. RWTD demonstrates broader improvement than the baselines: it is the only method to improve all five preference metrics and the overall GenEval score relative to the base model. While FAV and DRaFT utilize reward gradients to achieve stronger HPSv2 optimization, they produce less balanced changes across held-out metrics, both decreasing CLIP relative to the base model. The out-of-domain GenEval result is particularly informative because none of these models receive direct compositional supervision during post-training: RWTD improves overall GenEval score to 0.75, while FAV, DrPO, and DRaFT decrease it to 0.72, 0.72, and 0.62, respectively. These results suggest that RWTD is less prone to reward-specific overfitting and better preserves the generator's semantic and compositional capabilities.

\begin{figure*}[!t]
    \centering

    \begin{subfigure}[t]{0.38\textwidth}
        \centering
        \includegraphics[width=\linewidth]{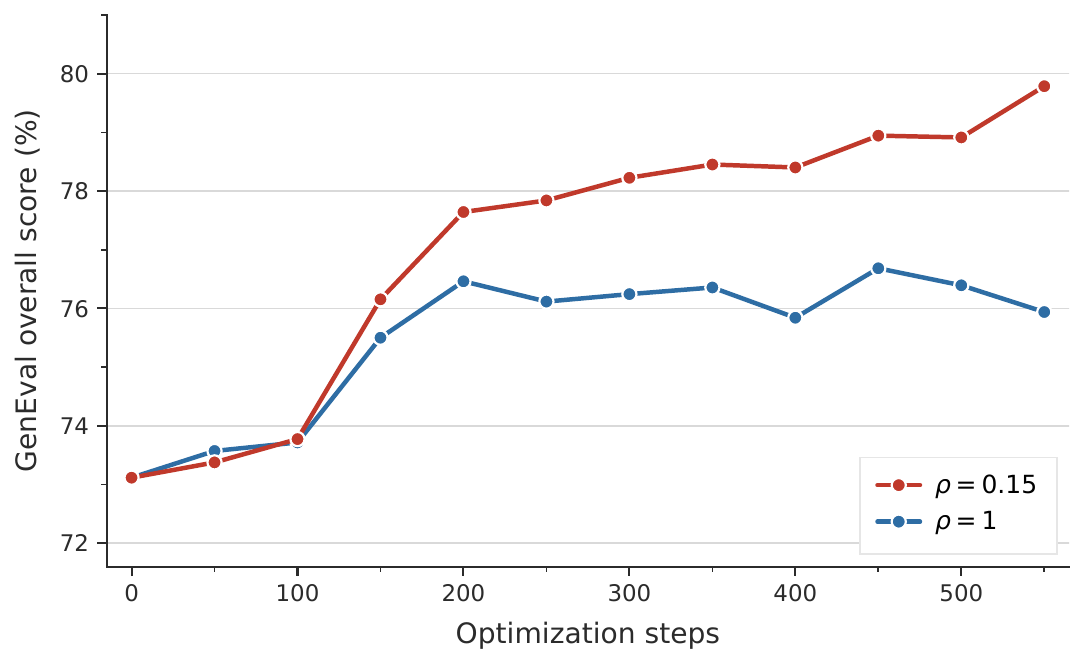}
        \caption{Head-to-head GenEval training comparison.}
        \label{fig:geneval-vs-steps}
    \end{subfigure}
    \hfill
    \begin{subfigure}[t]{0.38\textwidth}
        \centering
        \includegraphics[width=\linewidth]{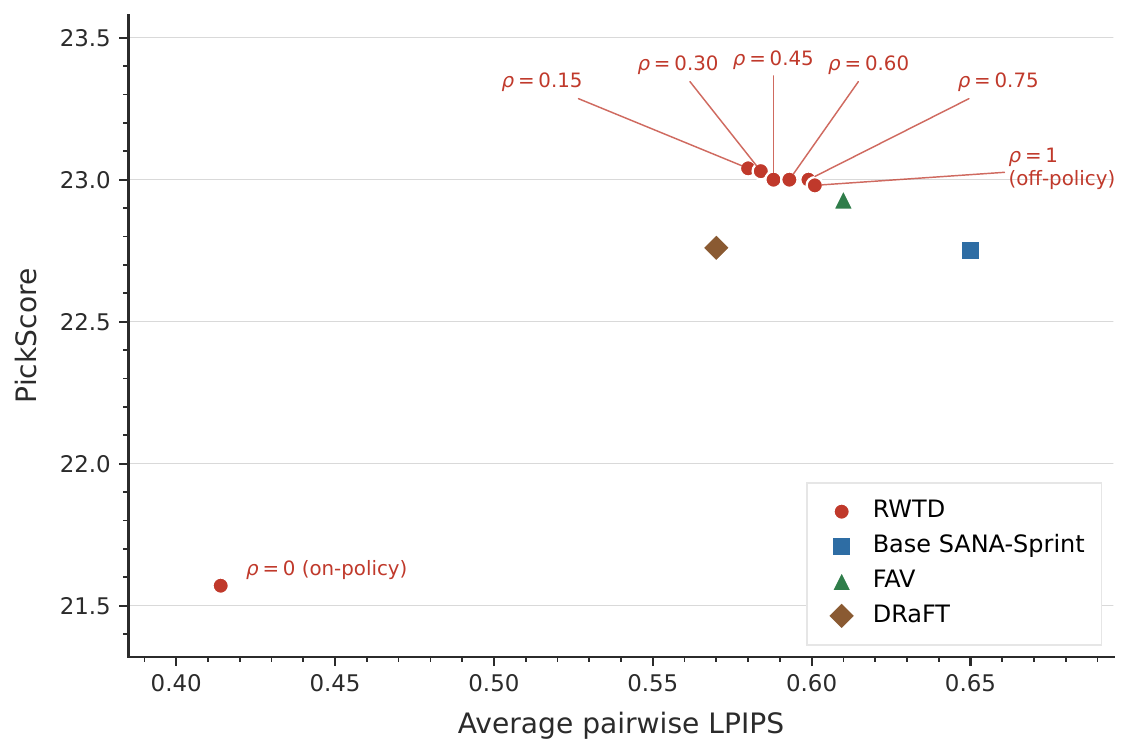}
        \caption{Held-out PickScore and diversity evaluation.}
        \label{fig:pickscore-vs-lpips}
    \end{subfigure}

    \caption{\textbf{Effect of mixed reward tilting.}
    \textbf{Left:} On GenEval, mixed RWTD ($\rho=0.15$) optimizes faster and reaches a substantially higher score than the canonical off-policy target ($\rho=1$).
    \textbf{Right:} Moderate reference mixing achieves the best held-out PickScore while retaining comparable perceptual diversity, whereas pure on-policy training ($\rho=0$) substantially reduces diversity.}
    \label{fig:rwtd-analysis}
    \vspace{-6pt}
\end{figure*}

\textbf{Mixed Targets Outperform Canonical Reference Tilting}
We examine whether incorporating current-model particles into the conventional reward-tilted reference target improves optimization on difficult tasks. Figure~\ref{fig:geneval-vs-steps} compares the conventional off-policy target ($\rho=1$) with mixed RWTD ($\rho=0.15$). Mixing accelerates optimization and reaches $0.80$ on GenEval, while off-policy RWTD plateaus near $0.77$, suggesting that current-model particles are beneficial in supplying high-reward targets when reference samples are weak. We thus view on-policy mixing as a way to improve reward optimization and data efficiency during training. 

Figure~\ref{fig:pickscore-vs-lpips} evaluates the reward--diversity trade-off using average pairwise LPIPS~\citep{zhang2018unreasonable} over twenty samples for 200 PartiPrompts. Moderate mixing achieves the highest held-out PickScore with comparable diversity to competing methods, whereas pure on-policy training degrades both, reflecting overly sharp updates. Thus, on-policy particles accelerate adaptation, while reference particles prevent premature concentration. Full results appear in Section~\ref{app:additional-results}.

% \subsection{Why is Optimal Transport Necessary?}
% \label{sec:transport-ablation}
\textbf{Why is Optimal Transport Necessary?} We isolate the role of OT by varying how per-sample targets are assigned, fixing all other hyperparameters. \textit{RWTD-Barycentric} is our default method and uses the conditional mean in Equation~\ref{eq:barycentric-target}. \textit{RWTD-Sampled} samples a target particle from each source particle's conditional coupling row. Finally, \textit{reward-weighted regression} (RWR) independently samples reward-weighted target particles, removing transport geometry while preserving the same target marginal. Table~\ref{tab:transport-ablation} contains full numerical results on PartiPrompts for all three variants.

\begin{wraptable}[9]{r}{0.42\columnwidth}
    \vspace{-0.8em}
    \setlength{\columnsep}{5pt}
    \centering
    \scriptsize
    \setlength{\tabcolsep}{3pt}
    \renewcommand{\arraystretch}{1.05}

    \begin{tabular}{lccc}
        \toprule
        Metric
        & Bary.
        & RWR
        & Sampled \\
        \midrule
        PS
        & \textbf{23.02}
        & 22.70
        & 22.98 \\
        HPSv2
        & 31.92
        & 31.39
        & \textbf{31.96} \\
        CLIP
        & 0.2768
        & 0.2733
        & \textbf{0.2769} \\
        Aes.
        & 6.751
        & \textbf{6.878}
        & 6.724 \\
        IR
        & \textbf{1.322}
        & 1.297
        & 1.316 \\
        \bottomrule
    \end{tabular}

    \caption{Transport-target ablation.}
    \label{tab:transport-ablation}
    \vspace{-0.9em}
\end{wraptable}

The barycentric and sampled variants perform similarly across all metrics, consistent with the barycentric target being the conditional expectation of the sampled target. In contrast, independent assignment RWR better improves Aesthetics but degrades metrics like CLIP and PickScore relative to the base model. This suggests that the primary benefit comes from the sample-specific assignments induced by feature-space OT, rather than from barycentric averaging itself, and that transport geometry promotes more balanced alignment and semantic preservation.

\section{Conclusion}

We introduced Reward-Weighted Transport Distillation, a general post-training method for one-step generators that requires only generated samples and scalar reward evaluations. Unlike canonical reference-only tilting, RWTD mixes tilted current and reference distributions, using current particles to accelerate adaptation while reference particles help preserve diversity. RWTD thus introduces a different view of one-step generator alignment: the alignment target need not remain fixed to a reward-tilted reference distribution, but can evolve with the improving generator while retaining an explicit reference anchor. This adaptive target yields distinct fixed-point behavior and empirically improves both reward optimization and semantic preservation relative to standard reference tilting.

%% file: sections/appendix.tex
\section{Proofs and Extended Theoretical Analysis}
\label{app:proofs}
% Complete fixed-point derivations, assumptions, and partial-transport proof.

\subsection{Derivation of the Mixed Reward-Tilting Fixed Point}
\label{app:mixed-fixed-point}

We prove Proposition~\ref{prop:mixed-fixed-point}. Define
\[
    s(\vx)=\exp(\beta r(\vx))
\]
and
\begin{equation}
    Z_{\mathrm{ref}}
    =
    \E_{\vx\sim p_{\mathrm{ref}}}
    \left[
        s(\vx)
    \right],
    \qquad
    q_{\mathrm{ref}}(\vx)
    =
    \frac{
        p_{\mathrm{ref}}(\vx)s(\vx)
    }{
        Z_{\mathrm{ref}}
    }
    =
    \mathcal{T}_{\beta,r}[p_{\mathrm{ref}}](\vx).
    \label{eq:app-reference-tilt}
\end{equation}

We first show that every reference-dominated fixed point induces a
solution of Equation~\ref{eq:lambda-characterization}. Let
$p_\infty\ll p_{\mathrm{ref}}$ be a fixed point of
$Q_{\beta,\rho}$ and define
\begin{equation}
    Z_\infty
    =
    \E_{\vx\sim p_\infty}
    \left[
        s(\vx)
    \right],
    \qquad
    \lambda
    =
    \frac{1-\rho}{Z_\infty}.
    \label{eq:app-lambda}
\end{equation}

The fixed-point condition is
\begin{equation}
    p_\infty(\vx)
    =
    (1-\rho)
    \frac{
        p_\infty(\vx)s(\vx)
    }{
        Z_\infty
    }
    +
    \rho q_{\mathrm{ref}}(\vx).
    \label{eq:app-population-fixed-point}
\end{equation}
Substituting Equation~\ref{eq:app-lambda} and collecting the terms
involving $p_\infty$ gives
\begin{equation}
    p_\infty(\vx)
    \left(
        1-\lambda s(\vx)
    \right)
    =
    \rho q_{\mathrm{ref}}(\vx).
    \label{eq:app-fixed-point-factorization}
\end{equation}

Because Equation~\ref{eq:app-population-fixed-point} implies
$p_\infty\geq\rho q_{\mathrm{ref}}$, the factor
$1-\lambda s(\vx)$ is positive
$q_{\mathrm{ref}}$-almost everywhere. Therefore,
\begin{equation}
    p_\infty(\vx)
    =
    \frac{
        \rho q_{\mathrm{ref}}(\vx)
    }{
        1-\lambda s(\vx)
    }
    =
    \frac{\rho}{Z_{\mathrm{ref}}}
    \frac{
        p_{\mathrm{ref}}(\vx)
        \exp\!\left(\beta r(\vx)\right)
    }{
        1-\lambda\exp\!\left(\beta r(\vx)\right)
    }.
    \label{eq:app-mixed-fixed-point}
\end{equation}

Normalizing Equation~\ref{eq:app-mixed-fixed-point} yields
\begin{equation}
    1
    =
    \rho
    \E_{\vx\sim q_{\mathrm{ref}}}
    \left[
        \frac{1}{1-\lambda s(\vx)}
    \right]
    =
    \rho H(\lambda).
\end{equation}
Hence
\[
    H(\lambda)=\frac{1}{\rho},
\]
so every reference-dominated fixed point induces a solution of
Equation~\ref{eq:lambda-characterization}.

We next show uniqueness. If $0\leq\lambda_1<\lambda_2$ are both in
the admissible domain of $H$, then, since $s(\vx)>0$,
\[
    \frac{1}{1-\lambda_1s(\vx)}
    <
    \frac{1}{1-\lambda_2s(\vx)}
\]
$q_{\mathrm{ref}}$-almost everywhere. Therefore, $H$ is strictly
increasing on its admissible domain, so
$H(\lambda)=1/\rho$ has at most one solution.

It remains to prove the converse. Suppose that an admissible
$\lambda\geq0$ satisfies
\[
    H(\lambda)=\frac{1}{\rho},
\]
and define
\begin{equation}
    p_\lambda(\vx)
    =
    \frac{
        \rho q_{\mathrm{ref}}(\vx)
    }{
        1-\lambda s(\vx)
    }.
    \label{eq:app-converse-fixed-point}
\end{equation}
The defining equation for $H$ ensures that $p_\lambda$ integrates to
one. Moreover,
\begin{align}
    \lambda
    \E_{\vx\sim p_\lambda}[s(\vx)]
    &=
    \rho
    \E_{\vx\sim q_{\mathrm{ref}}}
    \left[
        \frac{\lambda s(\vx)}
        {1-\lambda s(\vx)}
    \right] \\
    &=
    \rho
    \left(
        H(\lambda)-1
    \right)
    =
    1-\rho.
    \label{eq:app-lambda-normalizer-identity}
\end{align}
Thus, writing
$Z_\lambda=\E_{\vx\sim p_\lambda}[s(\vx)]$, we have
$\lambda Z_\lambda=1-\rho$. Combining this identity with
Equation~\ref{eq:app-converse-fixed-point} gives
\begin{align}
    Q_{\beta,\rho}[p_\lambda](\vx)
    &=
    (1-\rho)
    \frac{
        p_\lambda(\vx)s(\vx)
    }{
        Z_\lambda
    }
    +
    \rho q_{\mathrm{ref}}(\vx) \\
    &=
    \lambda s(\vx)p_\lambda(\vx)
    +
    \rho q_{\mathrm{ref}}(\vx) \\
    &=
    p_\lambda(\vx).
\end{align}
Therefore, every admissible solution of
Equation~\ref{eq:lambda-characterization} defines a
reference-dominated fixed point. Together with the strict
monotonicity of $H$, this proves the claimed one-to-one
correspondence and uniqueness.

Finally, when $\rho=1$, Equation~\ref{eq:lambda-characterization}
becomes $H(\lambda)=1$. Since $H(0)=1$ and $H$ is strictly
increasing, the unique solution is $\lambda=0$, yielding
\[
    p_\infty=q_{\mathrm{ref}}
    =
    \mathcal{T}_{\beta,r}[p_{\mathrm{ref}}].
\]
\hfill$\square$

\subsection{Existence, Uniqueness, and Effect of the Mixing Coefficient}
\label{app:rho-analysis}

Proposition~\ref{prop:mixed-fixed-point} shows that
reference-dominated fixed points are in one-to-one correspondence with
solutions of
\[
    H(\lambda)=\frac{1}{\rho},
\]
and that such a solution is unique whenever it exists. We now give a
sufficient condition ensuring existence for every $0<\rho\leq1$ and
analyze how the resulting fixed point varies with $\rho$.

We continue to restrict attention to distributions absolutely continuous
with respect to $p_{\mathrm{ref}}$. This is a natural invariant class:
if $p_0\ll p_{\mathrm{ref}}$, then
$Q_{\beta,\rho}[p_t]\ll p_{\mathrm{ref}}$ for every finite $t$.

Assume that $s(\vx)=\exp(\beta r(\vx))$ is essentially bounded under
$q_{\mathrm{ref}}$, and define
\begin{equation}
    s_{\max}
    =
    \operatorname*{ess\,sup}_{\vx\sim q_{\mathrm{ref}}}
    s(\vx).
\end{equation}

Recall that
\begin{equation}
    H(\lambda)
    =
    \E_{\vx\sim q_{\mathrm{ref}}}
    \left[
        \frac{1}{1-\lambda s(\vx)}
    \right],
    \qquad
    0\leq\lambda<s_{\max}^{-1}.
    \label{eq:app-lambda-normalization}
\end{equation}
On this interval, $H$ is continuous and strictly increasing, with
\begin{equation}
    H'(\lambda)
    =
    \E_{\vx\sim q_{\mathrm{ref}}}
    \left[
        \frac{
            s(\vx)
        }{
            \left(1-\lambda s(\vx)\right)^2
        }
    \right]
    >
    0.
    \label{eq:app-H-derivative}
\end{equation}

\textbf{Existence.}

Assume additionally that
\begin{equation}
    \lim_{\lambda\uparrow s_{\max}^{-1}}
    H(\lambda)
    =
    \infty.
    \label{eq:app-H-boundary}
\end{equation}
Since $H(0)=1$, continuity and strict monotonicity imply that $H$ maps
$[0,s_{\max}^{-1})$ bijectively onto $[1,\infty)$. Therefore, for every
$0<\rho\leq1$, there exists a unique
$\lambda_\rho\in[0,s_{\max}^{-1})$ satisfying
\begin{equation}
    H(\lambda_\rho)
    =
    \frac{1}{\rho}.
\end{equation}
For $0<\rho<1$, we have $\lambda_\rho>0$, while for $\rho=1$,
$\lambda_\rho=0$.

By Proposition~\ref{prop:mixed-fixed-point}, this solution defines the
unique reference-dominated fixed point
\begin{equation}
    p_\rho(\vx)
    =
    \frac{
        \rho q_{\mathrm{ref}}(\vx)
    }{
        1-\lambda_\rho s(\vx)
    }.
    \label{eq:app-constructed-fixed-point}
\end{equation}

The boundary condition in Equation~\ref{eq:app-H-boundary} holds
automatically when $\mathcal{X}$ is finite and
$p_{\mathrm{ref}}(\vx)>0$ for every $\vx\in\mathcal{X}$. Indeed,
$q_{\mathrm{ref}}$ assigns positive mass to a state attaining
$s_{\max}$, whose contribution to $H(\lambda)$ diverges as
$\lambda\uparrow s_{\max}^{-1}$.

\textbf{How does $\rho$ sharpen the tilt?}

We see from Equations \ref{eq:app-lambda-normalization} and \ref{eq:app-H-derivative} that $\lambda$ is a strictly decreasing function of $\rho$.
Indeed, implicit differentiation yields

\begin{equation}
    \frac{d\lambda}{d\rho}
    =
    -
    \frac{
        1
    }{
        \rho^2 H'(\lambda)
    }
    <
    0.
    \label{eq:app-lambda-rho-derivative}
\end{equation}

Thus, decreasing $\rho$ increases $\lambda$. At the off-policy endpoint
$\rho=1$, we have $\lambda=0$ and recover ordinary reference tilting. If
$H(\lambda)$ diverges as $\lambda\uparrow s_{\max}^{-1}$, then

\begin{equation}
    \lim_{\rho\downarrow0}\lambda(\rho)
    =
    s_{\max}^{-1}.
\end{equation}

Finally, increasing $\lambda$ sharpens the fixed point toward high-reward
samples. To see this, consider two samples $\vx_1$ and $\vx_2$ with
$q_{\mathrm{ref}}(\vx_i)>0$ and
$r(\vx_1)>r(\vx_2)$, and write $s_i=s(\vx_i)$. Then
\begin{equation}
    \frac{
        p_\rho(\vx_1)/q_{\mathrm{ref}}(\vx_1)
    }{
        p_\rho(\vx_2)/q_{\mathrm{ref}}(\vx_2)
    }
    =
    \frac{
        1-\lambda_\rho s_2
    }{
        1-\lambda_\rho s_1
    }.
\end{equation}

The logarithmic derivative of this ratio is

\begin{equation}
    \frac{d}{d\lambda}
    \log
    \left(
        \frac{1-\lambda s_2}
        {1-\lambda s_1}
    \right)
    =
    \frac{
        s_1-s_2
    }{
        (1-\lambda s_1)(1-\lambda s_2)
    }
    >
    0.
\end{equation}

Therefore, as $\rho$ decreases and $\lambda$ increases, the fixed point
assigns progressively greater relative mass to higher-reward samples.
The mixing coefficient $\rho$ thus interpolates between ordinary
reference reward tilting at $\rho=1$ and increasingly sharp on-policy
reward amplification as $\rho$ approaches zero.

\subsection{Proof of Fixed-Point Invariance}
\label{app:partial-fixed-point-invariance}

We prove Proposition~\ref{prop:partial-fixed-point-invariance}. For a
distribution $p$, let

\begin{equation}
    q_p
    =
    Q_{\beta,\rho}[p],
\end{equation}

and let $T_p$ be the exact quadratic-cost optimal transport map from $p$
to $q_p$, so that $(T_p)_\#p=q_p$. Define the interpolating map

\begin{equation}
    S_{p,\eta}(\vx)
    =
    (1-\eta)\vx+\eta T_p(\vx),
\end{equation}

such that the partial update is
$\mathcal{F}_\eta[p]=(S_{p,\eta})_\#p$.

Let

\begin{equation}
    D
    =
    W_2(p,q_p)
    =
    \left(
        \E_{\vx\sim p}
        \left[
            \left\|
                T_p(\vx)-\vx
            \right\|_2^2
        \right]
    \right)^{1/2}.
    \label{eq:app-full-transport-distance}
\end{equation}

The joint distribution of
$(\vx,S_{p,\eta}(\vx))$ for $\vx\sim p$ is a valid coupling between
$p$ and $\mathcal{F}_\eta[p]$. Therefore,

\begin{align}
    W_2\!\left(
        p,\mathcal{F}_\eta[p]
    \right)
    &\leq
    \left(
        \E_{\vx\sim p}
        \left[
            \left\|
                S_{p,\eta}(\vx)-\vx
            \right\|_2^2
        \right]
    \right)^{1/2}
    \\
    &=
    \eta D.
    \label{eq:app-partial-upper-bound}
\end{align}

Similarly, the joint distribution of
$(S_{p,\eta}(\vx),T_p(\vx))$ is a valid coupling between
$\mathcal{F}_\eta[p]$ and $q_p$, which gives

\begin{equation}
    W_2\!\left(
        \mathcal{F}_\eta[p],q_p
    \right)
    \leq
    (1-\eta)D.
    \label{eq:app-remaining-upper-bound}
\end{equation}

Applying the triangle inequality for $W_2$ and combining
Equations~\ref{eq:app-partial-upper-bound} and
\ref{eq:app-remaining-upper-bound}, we obtain

\begin{align}
    D
    &=
    W_2(p,q_p)
    \\
    &\leq
    W_2\!\left(
        p,\mathcal{F}_\eta[p]
    \right)
    +
    W_2\!\left(
        \mathcal{F}_\eta[p],q_p
    \right)
    \\
    &\leq
    \eta D+(1-\eta)D
    =
    D.
\end{align}

All inequalities must therefore be equalities. In particular,

\begin{equation}
    W_2\!\left(
        p,\mathcal{F}_\eta[p]
    \right)
    =
    \eta W_2(p,q_p).
    \label{eq:app-constant-speed-geodesic}
\end{equation}

We now prove the two directions of
Equation~\ref{eq:partial-fixed-point-equivalence}.

First, suppose that $\mathcal{F}_\eta[p]=p$ for some $\eta>0$. Then
Equation~\ref{eq:app-constant-speed-geodesic} implies

\begin{equation}
    0
    =
    W_2\!\left(
        p,\mathcal{F}_\eta[p]
    \right)
    =
    \eta W_2(p,q_p).
\end{equation}

Since $\eta>0$, we must have $W_2(p,q_p)=0$, and hence
$p=q_p=Q_{\beta,\rho}[p]$.

Conversely, suppose that $Q_{\beta,\rho}[p]=p$. Then $q_p=p$, so the
optimal quadratic transport cost is zero. Consequently,
$T_p(\vx)=\vx$ for $p$-almost every $\vx$, and therefore

\begin{equation}
    S_{p,\eta}(\vx)
    =
    (1-\eta)\vx+\eta T_p(\vx)
    =
    \vx
\end{equation}

for $p$-almost every $\vx$. It follows that
$\mathcal{F}_\eta[p]=p$ for every $0<\eta\leq1$.

Thus,

\begin{equation}
    \mathcal{F}_\eta[p]=p
    \quad\Longleftrightarrow\quad
    Q_{\beta,\rho}[p]=p
\end{equation}

for every $0<\eta\leq1$, proving that the fixed-point set is independent
of the partial transport step size. The exclusion of $\eta=0$ is
necessary because $\mathcal{F}_0[p]=p$ for every distribution $p$.
\hfill$\square$

The invariance of the idealized fixed point to the partial transport step size thus motivates the usage of partial transport as a means to ensure stable training with conservative updates, without changing the target distribution. 

\section{Extended Experimental Details}
\label{app:experimental-details}
% Hyperparameters, prompt construction, compute and reward-call accounting.

\subsection{Hyperparameters}
\label{sec:hyperparameters}

Detailed hyperparameter configurations are presented in Table \ref{tab:training-hyperparameters}. For each batch element, the Sinkhorn regularization is chosen adaptively as
\[
\varepsilon
=
\max\!\left(
s_{\mathrm{OT}}\,\operatorname{median}_{i,j} C_{ij},
\varepsilon_{\min}
\right),
\]
where \(C_{ij}\) is the weighted feature-space transport cost, \(s_{\mathrm{OT}}\) is the configured regularization scale, and \(\varepsilon_{\min}\) prevents numerical instability. In practice, we found that this led to stable Sinkhorn iterations, with the regularization parameter usually falling in the range $0.03 \leq \epsilon \leq 0.05$.

\begin{table*}[t]
    \centering
    \small
    \renewcommand{\arraystretch}{1.08}

    % ---------------------------------------------------------
    % SDXL-Turbo
    % ---------------------------------------------------------
    \begin{minipage}[t]{0.42\textwidth}
        \centering
        \textbf{(a) SDXL-Turbo}
        \vspace{3pt}

        \begin{tabular}{@{}p{0.70\linewidth}p{0.24\linewidth}@{}}
            \toprule
            Hyperparameter & GenEval post-training \\
            \midrule
            \multicolumn{2}{@{}l}{\textbf{Optimization}} \\
            LoRA rank                         & 16 \\
            LoRA $\alpha$                     & 16 \\
            AdamW weight decay                & 0.0 \\
            AdamW $\beta_1$                   & 0.9\\
            AdamW $\beta_2$                   & 0.999\\
            Learning rate                     & 5e-5\\
            Number of GPUs                    & 4\\
            Gradient accumulation steps       & 8\\
            Per-GPU batch size                & 1\\
            Number of optimization steps      & 300\\
            \midrule
            \multicolumn{2}{@{}l}{\textbf{RWTD}} \\
            Reward temperature $\beta$        & 4.0 \\
            Reference mass $\rho$             & 0.15 \\
            Partial transport step $\eta$     & 0.20 \\
            Current particles $N$             & 24 \\
            Reference particles $M$           & 24 \\
            Feature encoder                   & MAE \\
            Entropic OT scale $s_\text{OT}$      & 0.1 \\
            Entropic OT minimum $\epsilon_\text{min}$     & 0.0001 \\
            Sinkhorn iterations $L_{\mathrm{SK}}$ & 100 \\
            \bottomrule
        \end{tabular}
    \end{minipage}
    \hfill
    % ---------------------------------------------------------
    % SANA Sprint 1.6B
    % ---------------------------------------------------------
    \begin{minipage}[t]{0.55\textwidth}
        \centering
        \textbf{(b) SANA Sprint 1.6B}
        \vspace{3pt}

        \begin{tabular}{@{}p{0.52\linewidth}
                            p{0.21\linewidth}
                            p{0.21\linewidth}@{}}
            \toprule
            Hyperparameter
            & \centering\arraybackslash
              \shortstack{GenEval\\post-training}
            & \centering\arraybackslash
              \shortstack{HPSv2\\post-training} \\
            \midrule
            \multicolumn{3}{@{}l}{\textbf{Optimization}} \\
            LoRA rank                         & 32 & 32 \\
            LoRA $\alpha$                     & 32 & 32 \\
            AdamW weight decay                & 0.0 & 0.0 \\
            AdamW $\beta_1$                   & 0.9 & 0.9 \\
            AdamW $\beta_2$                   & 0.999 & 0.999 \\
            Learning rate                     & 3e-5 & 1e-4 \\
            Number of GPUs                    & 8 & 8 \\
            Gradient accumulation steps       & 4 & 4 \\
            Per-GPU batch size                & 1 & 1 \\
            Number of optimization steps      & 550 & 400 \\
            \midrule
            \multicolumn{3}{@{}l}{\textbf{RWTD}} \\
            Reward temperature $\beta$        & 2.0 & {5.0} \\
            Reference mass $\rho$             & 0.15 & {0.15} \\
            Partial transport step $\eta$     & 0.20 & {0.20} \\
            Current particles $N$             & 24 & 24 \\
            Reference particles $M$           & 24 & 24 \\
            Feature encoder                   & DINOv2 & DINOv2 \\
            Entropic OT scale $s_\text{OT}$     & 0.1 & 0.1 \\
            Entropic OT minimum $\epsilon_\text{min}$     & 0.0001 & 0.0001 \\
            Sinkhorn iterations $L_{\mathrm{SK}}$ & 100 & 100 \\
            \bottomrule
        \end{tabular}
    \end{minipage}

    \caption{
        Optimization and RWTD hyperparameters for SDXL-Turbo and
        SANA Sprint 1.6B. SANA settings are reported separately for
        GenEval and HPSv2 post-training.
    }
    \label{tab:training-hyperparameters}
\end{table*}

\begin{table*}[t]
    \centering
    \small
    \setlength{\tabcolsep}{6pt}
    \renewcommand{\arraystretch}{1.08}

    \begin{tabular}{llcccccc}
        \toprule
        Reward
        & Method
        & GPUs
        & \shortstack{Effective\\batch \\(particles)}
        & \shortstack{Reward calls\\per step}
        & \shortstack{Reward-gradient\\backprop.}
        & Steps
        & GPU-hours \\
        \midrule

        \multirow{3}{*}{GenEval}
        & RWTD & 8 & 1536 & 1536 & No & 550 & 197 \\
        & DrPO & 8 & 1536 & 768  & No & 600 & 194 \\
        & FAV  & 8 & 1536 & 6144 & No & 550 & 431 \\

        \midrule

        \multirow{3}{*}{HPSv2}
        & RWTD & 8 & 1536 & 1536 & No  & 400 & 121 \\
        & DrPO & 8 & 1536 & 768  & No  & 400 & 119 \\
        & FAV  & 8 & 1536 & 768  & Yes & 400 & 123 \\
        & DRaFT & 8 & 1536 & 1536 & Yes & 400 & 315 \\

        \bottomrule
    \end{tabular}

    \caption{
        Computational cost of post-training SANA Sprint 1.6B. All methods
        use identical hardware and effective batch sizes. Reward calls count
        scalar reward evaluations, while reward-gradient backpropagation
        indicates whether gradients are propagated through the reward model.
        GPU-hours measure total wall clock runtime multiplied by the number of GPUs.
    }
    \label{tab:computational-cost}
\end{table*}

\subsection{Computational Cost and Training Efficiency}
\label{sec:computational-cost}

Table~\ref{tab:computational-cost} compares the end-to-end computational cost of post-training SANA Sprint 1.6B under identical hardware and effective particle batch sizes. On GenEval, RWTD requires 197 GPU-hours, comparable to the 194 GPU-hours required by DrPO, despite using twice as many scalar reward evaluations per step. In contrast, FAV requires 431 GPU-hours, approximately $2.2\times$ the cost of RWTD. This increase reflects the additional reward evaluations required by its zeroth-order estimator for the non-differentiable GenEval reward.

The costs are generally close under HPSv2 post-training: RWTD requires 121 GPU-hours, compared with 119 for DrPO, 123 for FAV, and 315 for DRaFT. Here, FAV and DRaFT can differentiate through the HPSv2 reward, therefore avoiding the multiple evaluations required for zeroth-order estimation, although incurring reward-gradient backpropagation overhead. RWTD remains gradient-free and incurs only a modest overhead over DrPO, suggesting that frozen feature encoding and Sinkhorn transport contribute relatively little to the total training cost at the particle sizes used in our experiments. Overall, RWTD provides its alignment improvements at a computational cost comparable to existing methods, while remaining applicable to non-differentiable rewards.

\subsection{Datasets and Models}
\label{sec:data-and-models}

In this section, we discuss the details of the datasets, reward models, and feature encoders used during training. 

\textbf{GenEval Training Prompts} We post-train SDXL Turbo and SANA Sprint 1.6B using 800 total GenEval prompts, generated using the official GenEval codebase (MIT License) ~\citep{ghosh2023geneval}. We filter out prompts that overlap with the evaluation set to ensure fair evaluation. For SDXL Turbo training, the prompt set consists of 160 color attribution, 160 color, 160 counting, 160 position, and 160 two-object prompts. For SANA Sprint 1.6B training, the prompt set consists of 280 color attribution, 80 color, 160 counting, 140 position, 140 two-object prompts. 

\textbf{Pick-a-pic v2} In the preference alignment experiments, we post-train using the Pick-a-pic v2 prompt set~\citep{kirstain2023pick}. We use a filtered SFW subset of 15485 safe-for-work, user-written text prompts drawn from the Pick-a-Pic image-preference dataset. It contains diverse, naturally occurring text-to-image requests and excludes prompts flagged as explicit or otherwise unsafe.

\textbf{GenEval training reward.}
We use a dense, detector-based adaptation of GenEval as the training reward. Each generated image is evaluated against structured prompt metadata specifying the required object classes, counts, colors, and spatial relations, as well as any forbidden extra objects. A Mask2Former detector identifies COCO objects, while an OpenCLIP classifier predicts the color of each detected object crop. Following the official GenEval protocol, an image is marked correct only if all required objects and attributes are present, no forbidden count is reached, and every spatial relation is satisfied.

To provide a more fine-grained learning signal than binary correctness alone, we compute partial-credit scores for each metadata clause. Object-presence credit is the mean confidence of the top required detections, with missing detections assigned zero credit. Absence credit is one minus the confidence of the first forbidden extra detection. Color credit is the OpenCLIP probability assigned to the requested color, and spatial credit varies continuously with the normalized directional margin between object centers. The dense score is the mean over all applicable clauses,
\[
r_{\mathrm{dense}}(\vx,c)
=
\frac{1}{|\mathcal{C}(c)|}
\sum_{j \in \mathcal{C}(c)} s_j(\vx,c),
\qquad s_j \in [0,1],
\]
where \(\vx\) is the generated image, \(c\) is its structured prompt metadata, and \(\mathcal{C}(c)\) denotes its presence, absence, color, and spatial-relation clauses. Our hybrid training reward additionally gives a bonus for satisfying the complete official GenEval criterion:
\[
r_{\mathrm{GenEval}}(\vx,c)
=
r_{\mathrm{dense}}(\vx,c)
+
k\,
\mathbf{1}\!\left[\operatorname{GenEvalCorrect}(\vx,c)\right],
\qquad k = 0.50.
\]
Thus, the dense component rewards incremental improvements toward satisfying individual constraints, while the binary bonus preserves the original benchmark objective of complete compositional correctness. The scorer is non-differentiable and is evaluated outside the model's computation graph. For fair comparison, we additionally trained all baselines with this dense reward.

\textbf{HPSv2 Training Reward} We use the official HPSv2 (Human Preference Score v2) reward model checkpoint \citep{wu2023human}. HPSv2 is a CLIP-style encoder trained from large-scale human judgments to score how well an image matches a prompt while reflecting human preferences for visual quality and aesthetics; higher scores indicate greater predicted human preference.

\textbf{Feature Encoders} RWTD and DrPO training rely on the use of a frozen feature encoder, following standard practices from previous work~\citep{deng2026generative, han2026one, feng2026representation}. For SDXL Turbo post-training, we use the MAE~\citep{he2022masked} encoder already used by the official DrPO codebase. For SANA Sprint 1.6B post-training, we use the DINOv2~\citep{oquab2023dinov2} encoder. Strong RWTD empirical results suggest that the objective can generalize well across both different one-step generator backbones and feature encoders. 

\subsection{Baseline Implementations}
\label{sec:baselines}

For the SDXL Turbo experiments, we used the official DrPO codebase and replaced its GenEval reward evaluator with our dense reward. This modification affected only reward computation; the DrPO objective and optimization procedure were otherwise unchanged. We trained DrPO for 800 optimizer steps, compared with 300 steps for RWTD, because shorter DrPO runs did not achieve competitive validation performance.

For SANA-Sprint-1.6B, no directly compatible implementations of DrPO and DRaFT were available, so we reimplemented them within the same training framework used for RWTD. For FAV, we used the official FAV codebase, adding our dense GenEval reward. All methods used the same pretrained SANA-Sprint-1.6B checkpoint, one-step generation configuration, and LoRA parameterization. We initialized method-specific hyperparameters from the corresponding papers or public implementations and performed tuning on held-out validation prompts. This tuning was restricted to method-specific quantities such as reward weighting, kernel temperatures, and update scales. 

\section{Additional Results}
\label{app:additional-results}
% Full GenEval categories, sweeps, qualitative samples, and diagnostics.

\subsection{Ablation Studies}
\label{sec:ablations}

\textbf{Hyperparameter Sweeps} We perform coarse hyperparameter sweeps for the inverse reward temperature $\beta$, reference mass $\rho$, and partial transport step size $\eta$. Results are shown in Table \ref{tab:beta-rho-eta-ablation}. 

Increasing the inverse reward temperature generally strengthens preference optimization: HPSv2, Aesthetics, and ImageReward improve as $\beta$ increases. However, this sharper reward weighting is accompanied by a gradual decline in CLIP score, while PickScore peaks at $\beta=5$. These results suggest a trade-off between aggressively concentrating on high-reward particles and preserving broader semantic alignment. We therefore use $\beta=5$ as a balanced default.

While pure on-policy training ($\rho = 0$) results in severe degradation of most metrics, reflecting the overly aggressive distribution sharpening updates, we generally see that smaller (but non-zero) reference mass results in better optimization of the training reward HPSv2. Cross-reward generalization is demonstrated for all values $\rho > 0$, with $\rho = 0.15$ achieving the best PickScore, Aesthetics, and ImageReward, while $\rho = 0.45$ achieves the strongest CLIP score. We thus select $\rho = 0.15$ for its strong balance between in-domain reward optimization and cross-reward generalization.  

Finally, we see that RWTD is relatively robust to the transport step size $\eta$, with PickScore, CLIP, and ImageReward peaking at $\eta = 0.2$, but with HPSv2 and Aesthetics peaking at larger $\eta$ values. We choose $\eta = 0.2$ to balance cross-reward generalization with in-domain reward optimization. 

\begin{table*}[t]
    \centering
    \small
    \renewcommand{\arraystretch}{1.08}

    % ---------------------------------------------------------
    % Beta sweep
    % ---------------------------------------------------------
    \begin{minipage}[t]{0.32\textwidth}
        \centering
        \textbf{(a) Reward temperature $\beta$}
        \vspace{3pt}

        \setlength{\tabcolsep}{5pt}
        \resizebox{\linewidth}{!}{%
        \begin{tabular}{cccccc}
            \toprule
            $\beta$
            & PS
            & HPSv2
            & CLIP
            & Aes.
            & IR \\
            \midrule

            $1$
            & 22.98
            & 31.52
            & \textbf{0.2769}
            & 6.696
            & 1.289 \\

            $2$
            & 23.02
            & 31.92
            & 0.2768
            & 6.751
            & 1.322 \\

            $5$
            & \textbf{23.04}
            & 32.45
            & 0.2761
            & 6.800
            & 1.358 \\

            $10$
            & 23.00
            & \textbf{32.69}
            & 0.2749
            & \textbf{6.824}
            & \textbf{1.366} \\

            \bottomrule
        \end{tabular}%
        }
    \end{minipage}
    \hfill
    % ---------------------------------------------------------
    % Rho sweep
    % ---------------------------------------------------------
    \begin{minipage}[t]{0.32\textwidth}
        \centering
        \textbf{(b) Reference mass $\rho$}
        \vspace{3pt}

        \setlength{\tabcolsep}{5pt}
        \resizebox{\linewidth}{!}{%
        \begin{tabular}{cccccc}
            \toprule
            $\rho$
            & PS
            & HPSv2
            & CLIP
            & Aes.
            & IR \\
            \midrule

            $0$
            & 21.57 & 28.50 & 0.2561 & \textbf{6.814} & 1.071 \\

            $0.15$
            & \textbf{23.04} & \textbf{32.45} & 0.2761 & {6.800} & \textbf{1.358} \\

            $0.30$
            & 23.03 & 32.05 & 0.2766 & 6.713 & 1.325 \\

            $0.45$
            & 23.00 & 31.74 & \textbf{0.2772} & 6.719 & 1.296 \\

            $0.60$
            & 23.00 & 31.72 & 0.2771 & 6.687 & 1.294 \\

            $0.75$
            & 23.00 & 31.64 & 0.2769 & 6.685 & 1.287 \\

            $1.0$
            & 22.98 & 31.52 & 0.2766 & 6.681 & 1.277 \\

            \bottomrule
        \end{tabular}%
        }
    \end{minipage}
    \hfill
    % ---------------------------------------------------------
    % Eta sweep
    % ---------------------------------------------------------
    \begin{minipage}[t]{0.32\textwidth}
        \centering
        \textbf{(c) Partial transport step $\eta$}
        \vspace{3pt}

        \setlength{\tabcolsep}{5pt}
        \resizebox{\linewidth}{!}{%
        \begin{tabular}{cccccc}
            \toprule
            $\eta$
            & PS
            & HPSv2
            & CLIP
            & Aes.
            & IR \\
            \midrule

            $0.2$
            & \textbf{23.04} & {32.45} & \textbf{0.2761} & {6.800} & \textbf{1.358} \\

            $0.4$
            & 23.03 & 32.48 & 0.2756 & 6.817 & 1.357 \\

            $0.6$
            & 23.02 & 32.47 & 0.2754 & \textbf{6.820} & 1.353 \\

            $0.8$
            & 23.01 & \textbf{32.51} & 0.2758 & 6.805 & 1.344 \\

            $1.0$
            & 23.03 & 32.45 & 0.2758 & 6.783 & 1.353 \\

            \bottomrule
        \end{tabular}%
        }
    \end{minipage}

    \caption{
        Ablations of the reward temperature $\beta$, reference mass $\rho$,
        and partial transport step size $\eta$ for HPSv2 post-training on
        PartiPrompts. PS, Aes., and IR denote PickScore, Aesthetics, and
        ImageReward, respectively. Bold denotes the best result in each column.
    }
    \label{tab:beta-rho-eta-ablation}
\end{table*}

% \paragraph{Why Is Optimal Transport Necessary?}
% \label{app:ot-ablation}

% Let $\pi_i(k)=P_{\eps,ik}/a_i$ denote the conditional target assignment induced by the OT coupling for source feature $\vu_i$. \textit{RWTD-Barycentric} uses its conditional mean, $\sum_k\pi_i(k)\vv_k$, as the transport destination. \textit{RWTD-Sampled} instead draws $K_i\sim\pi_i$ and uses the single target feature $\vv_{K_i}$; its destination therefore equals the barycentric target in expectation. \textit{RWR-Independent} draws $K_i\sim\vb$ independently of $\vu_i$, preserving the reward-weighted target marginal while removing the source-conditioned OT assignment. Each destination is interpolated with $\vu_i$ using the same partial step $\eta$ from Equation ~\ref{eq:barycentric-target}.

\textbf{Sample Diversity}

We additionally measure the diversity of images generated by our model by computing the average pairwise LPIPS~\citep{zhang2018unreasonable} distance between images per prompt. We use 200 prompts from the PartiPrompts set and generate 20 images per prompt, resulting in 190 pairwise comparisons per prompt. 

Table~\ref{tab:lpips-diversity} shows that purely on-policy RWTD ($\rho=0$) substantially reduces perceptual diversity, reflecting the recursive reward tilting that concentrates the generator distribution. Introducing even modest reference mass ($\rho=0.15$) increases average pairwise LPIPS from $0.414$ to $0.580$, and the evaluated mixed configurations remain stable around $0.58$--$0.60$. These values are in the same range as competing methods, despite RWTD's sharper mixed reward-tilted target. Although all aligned models remain below the base model ($0.652$), the results suggest that reference mixing substantially mitigates the diversity loss of pure on-policy tilting while retaining RWTD's reward-alignment benefits.

\begin{table*}[t]
    \centering
    \setlength{\tabcolsep}{4pt}
    \renewcommand{\arraystretch}{1.12}
    \scriptsize

    \begin{tabular}{lcccccccccc}
        \toprule
        Metric
        & \shortstack{SANA Sprint\\1.6B}
        & DRaFT
        & FAV
        & \shortstack{RWTD\\$\rho=0$}
        & \shortstack{RWTD\\$\rho=0.15$}
        & \shortstack{RWTD\\$\rho=0.30$}
        & \shortstack{RWTD\\$\rho=0.45$}
        & \shortstack{RWTD\\$\rho=0.60$}
        & \shortstack{RWTD\\$\rho=0.75$}
        & \shortstack{RWTD\\$\rho=1.0$} \\
        \midrule

        \shortstack[l]{Avg. pairwise\\LPIPS $\uparrow$}
        & \textbf{0.652}
        & 0.571
        & 0.611
        & 0.414
        & 0.580
        & 0.584
        & 0.588
        & 0.593
        & 0.599
        & 0.601 \\

        \bottomrule
    \end{tabular}

    \caption{
        Sample diversity measured by the average pairwise LPIPS distance
        between samples generated for the same prompt. Higher LPIPS indicates
        greater perceptual diversity.
    }
    \label{tab:lpips-diversity}
\end{table*}

\subsection{Does Early Stopping Benefit Gradient-based Baselines?}
\label{sec:early_stopping}

We evaluate the earliest saved checkpoints of DRaFT and FAV (step 100)
to assess whether stopping earlier mitigates the semantic degradation
observed after HPSv2 post-training. Table~\ref{tab:early_stopping}
compares these checkpoints with their final checkpoints, using the
pretrained model and final RWTD model as references.

Both gradient-based baselines already exhibit reduced semantic alignment at step 100:
HPSv2 increases substantially, while CLIP falls below the pretrained
model. Earlier stopping partially mitigates DRaFT's CLIP degradation,
whereas FAV's final checkpoint improves all reported metrics over its
step-100 checkpoint. Neither early checkpoint recovers RWTD's
combination of PickScore, CLIP, and ImageReward, despite achieving
higher HPSv2. Thus, semantic degradation in these baselines is already
present early in training, rather than arising only after prolonged
optimization.

\begin{table}[t]
    \centering
    % \small
    \setlength{\tabcolsep}{5pt}
    \renewcommand{\arraystretch}{1.3}
    \begin{tabular}{lccccc}
        \hline
        Method & PickScore & HPSv2 (in-domain) & CLIP & Aesthetics & ImageReward \\
        \hline
        SANA Sprint 1.6B & 22.75 & 30.31 & 0.2753 & 6.565 & 1.148 \\
        RWTD (final) & \textbf{23.04} & 32.45 &
        \textbf{0.2761} & 6.800 & \textbf{1.358} \\
        \hline
        DRaFT (step 100) & 22.70 & 34.87 & 0.2655 & 7.089 & 1.291 \\
        DRaFT (step 400) & 22.76 & \textbf{37.02} &
        0.2573 & \textbf{7.252} & 1.312 \\
        \hline
        FAV (step 100) & 22.81 & 34.21 & 0.2711 & 6.844 & 1.281 \\
        FAV (step 400) & 22.93 & 35.74 & 0.2718 & 6.949 & 1.322 \\
        \hline
    \end{tabular}
    \caption{Early and final checkpoints of gradient-based baselines
    under HPSv2 post-training, evaluated on PartiPrompts. Step 100 is
    the earliest saved checkpoint for both baselines. Higher is better
    for all metrics; bold denotes the best value in each column.}
    \label{tab:early_stopping}
\end{table}

% \subsection{Additional Qualitative Results}

% We provide additional visual comparisons between RWTD models and baselines.

% \begin{figure}[t]
%     \centering

    \begin{minipage}[t]{0.63\textwidth}
        \footnotesize
        \begin{ruledalgorithm}{Reward-Weighted Transport Distillation}
            \label{alg:rwtd}

            \begin{algorithmic}[1]
                \Require Current generator $G_\theta$, reference generator
                $G_{\mathrm{ref}}$, reward $r$, frozen encoder $\phi$
                \Require Distributions $p_z,p_{\mathrm{prompt}}$; parameters
                $\beta,\rho,\eta,\eps$; particle counts $N,M$
                \Require Training steps $T$, Sinkhorn iterations
                $L_{\mathrm{SK}}$, learning rate $\lr$

                \For{$t=1,\ldots,T$}
                    \State Sample $c\sim p_{\mathrm{prompt}}$,
                    $\{\vz_i\}_{i=1}^{N}\sim p_z$,
                    and $\{\vz'_j\}_{j=1}^{M}\sim p_z$
                    \State $\vx_i\gets G_\theta(\vz_i,c)$,
                    \quad $i=1,\ldots,N$
                    \State $\vy_j\gets G_{\mathrm{ref}}(\vz'_j,c)$,
                    \quad $j=1,\ldots,M$

                    \State $\vw^\theta
                    \gets
                    \softmax_i\!\left(\beta r(\vx_i,c)\right)$
                    \State $\vw^{\mathrm{ref}}
                    \gets
                    \softmax_j\!\left(\beta r(\vy_j,c)\right)$

                    \State $\vu_i\gets\phi(\vx_i)$,
                    \quad $i=1,\ldots,N$
                    \State $\{\vv_k\}_{k=1}^{N+M}
                    \gets
                    \{\vu_i\}_{i=1}^{N}
                    \mathbin{\|}
                    \{\phi(\vy_j)\}_{j=1}^{M}$

                    \State $\va\gets\frac{1}{N}\vone_N$
                    \State $\vb\gets
                    \left[
                        (1-\rho)\vw^\theta;
                        \rho\vw^{\mathrm{ref}}
                    \right]$
                    \State $\mC_{ik}\gets
                    \left\|\vu_i-\vv_k\right\|_2^2$
                    for all $i,k$

                    \State $\mP_\eps\gets
                    \Call{Sinkhorn}
                    {\mC,\va,\vb,\eps,L_{\mathrm{SK}}}$

                    \State $\bar{\vu}_i\gets
                    \displaystyle
                    \frac{1}{a_i}
                    \sum_{k=1}^{N+M}
                    P_{\eps,ik}\vv_k$,
                    \quad $i=1,\ldots,N$
                    \State $\widetilde{\vu}_i\gets
                    (1-\eta)\vu_i+\eta\bar{\vu}_i$,
                    \quad $i=1,\ldots,N$

                    \State $\Ls_{\mathrm{RWTD}}\gets
                    \displaystyle
                    \frac{1}{N}
                    \sum_{i=1}^{N}
                    \left\|
                        \phi\!\left(G_\theta(\vz_i,c)\right)
                        -
                        \operatorname{sg}\!\left(\widetilde{\vu}_i\right)
                    \right\|_2^2$
                    \State $\theta\gets
                    \theta-\lr\nabla_\theta\Ls_{\mathrm{RWTD}}$
                \EndFor

                \State \Return $\theta$
            \end{algorithmic}
        \end{ruledalgorithm}
    \end{minipage}
    \hfill
    \begin{minipage}[t]{0.34\textwidth}
        \footnotesize
        \begin{ruledalgorithm}{Sinkhorn}
            \label{alg:sinkhorn}

            \begin{algorithmic}[1]
                \Require Cost matrix $\mC\in\R^{N\times K}$
                \Require Marginals $\va\in\R_+^N$ and
                $\vb\in\R_+^K$
                \Require Regularization $\eps>0$ and
                iterations $L_{\mathrm{SK}}$

                \State $\mK\gets
                \exp\!\left(-\mC/\eps\right)$
                \Comment{Gibbs kernel}
                \State $\vt\gets\vone_K$
                \Comment{Initialize}

                \For{$\ell=1,\ldots,L_{\mathrm{SK}}$}
                    \State $\vs\gets
                    \va\oslash(\mK\vt)$
                    \State $\vt\gets
                    \vb\oslash(\mK^\top\vs)$
                \EndFor

                \State $\mP_\eps\gets
                \operatorname{diag}(\vs)\,
                \mK\,
                \operatorname{diag}(\vt)$
                \Comment{Coupling}

                \State \Return $\mP_\eps$
                \Statex
                \textit{$\exp$ and $\oslash$ are elementwise.}
            \end{algorithmic}
        \end{ruledalgorithm}
    \end{minipage}
% \end{figure}

\begin{figure*}[!b]
    \centering

    \begin{subfigure}[t]{0.72\textwidth}
        \centering
        \includegraphics[width=\linewidth]{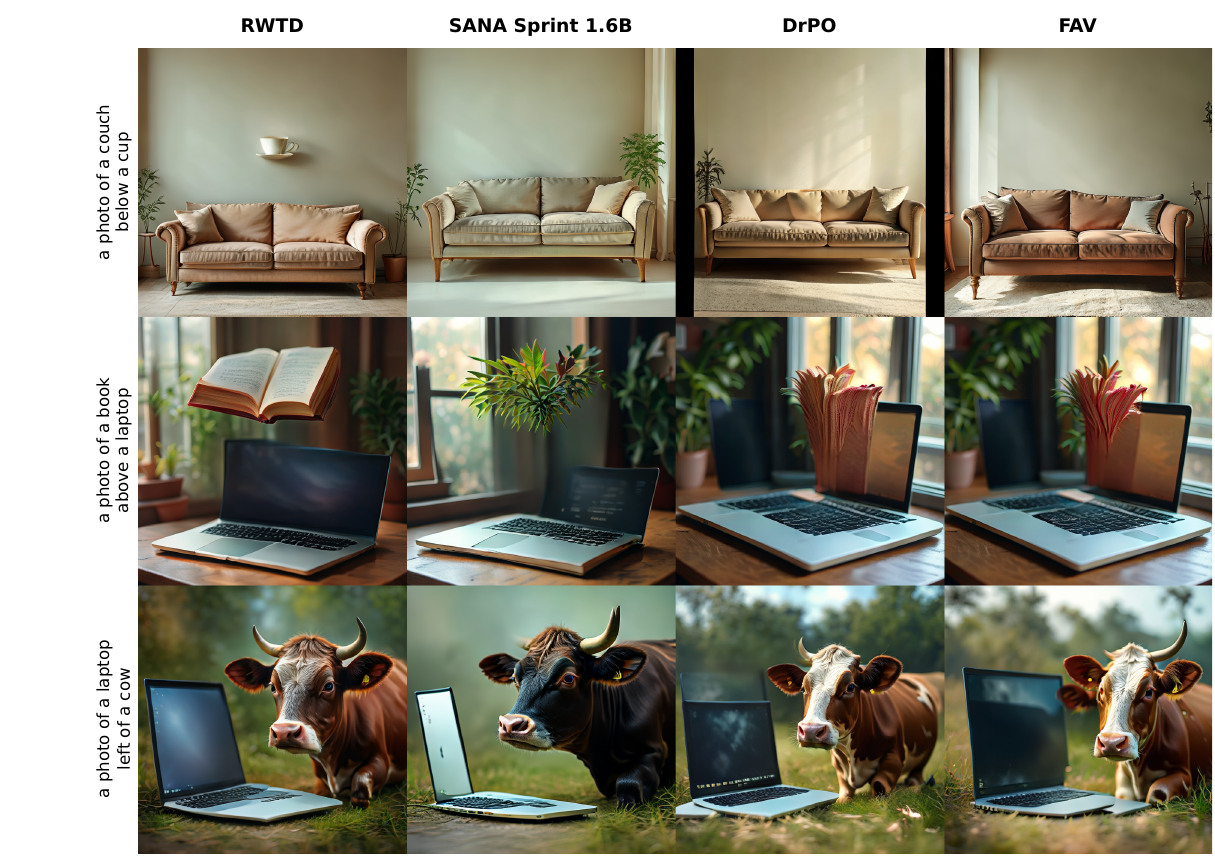}
        \caption{
            SANA Sprint 1.6B comparisons. RWTD improves alignment on difficult
            position prompts, while baselines omit the required second object
            or introduce visual artifacts.
        }
        \label{fig:sana-geneval}
    \end{subfigure}

    \vspace{0.2em}

    \begin{subfigure}[t]{0.66\textwidth}
        \centering
        \includegraphics[width=\linewidth]{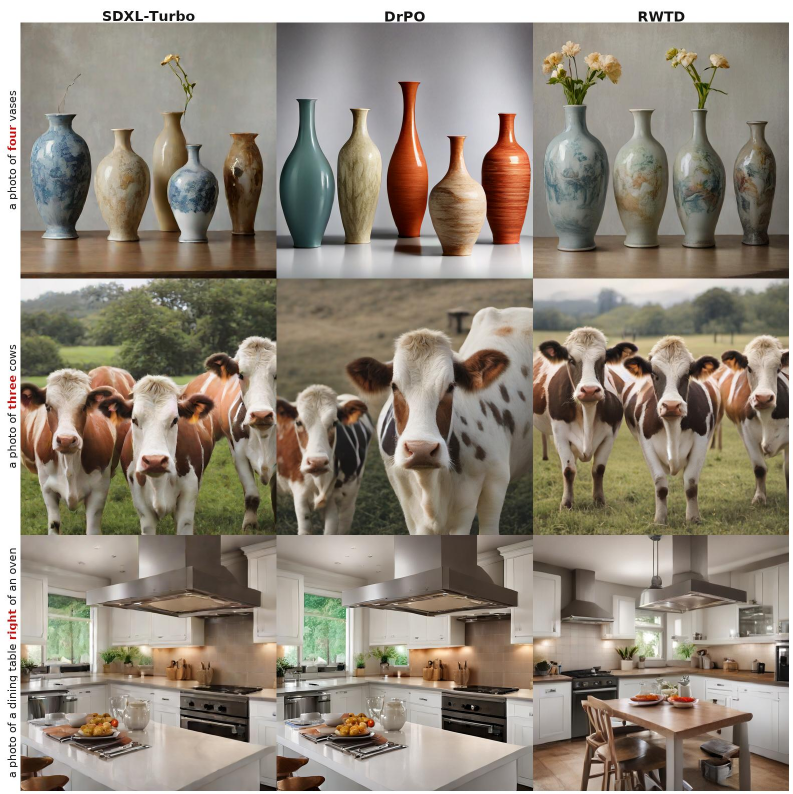}
        \caption{
            SDXL-Turbo comparisons. RWTD correctly satisfies the requested
            counts of four vases and three cows, and places the table to the
            right of the oven.
        }
        \label{fig:sdxl-turbo-geneval}
    \end{subfigure}

    \caption{
        Qualitative GenEval results showing improved counting and spatial
        composition after post-training with RWTD.
    }
    \label{fig:geneval-qualitative}

    \vspace{-6pt}
\end{figure*}